\documentclass{article}

\usepackage{arxiv}
\usepackage[numbers]{natbib}
\usepackage[utf8]{inputenc} % allow utf-8 input
\usepackage[T1]{fontenc}    % use 8-bit T1 fonts
\usepackage[dvipsnames]{xcolor}
\usepackage[hidelinks]{hyperref}       % hyperlinks
\hypersetup{
    colorlinks=true,
    linkcolor=MidnightBlue,
    citecolor=MidnightBlue,
    urlcolor=MidnightBlue
}
\usepackage{url}            % simple URL typesetting
\usepackage{booktabs}       % professional-quality tables
\usepackage{amsfonts}       % blackboard math symbols
\usepackage{nicefrac}       % compact symbols for 1/2, etc.
\usepackage{microtype}      % microtypography
\usepackage{lipsum}		% Can be removed after putting your text content
\usepackage{graphicx}
\usepackage{doi}
\usepackage{tikz}
\usepackage{amsmath}
\usepackage{tikz-cd}
\usepackage{longtable}
\usepackage{multirow}
\usepackage{svg}
\usepackage{float}
\usetikzlibrary{positioning, arrows.meta, calc, shadows.blur, fit, backgrounds}
\usepackage{colortbl}   % cell backgrounds in tables
\usepackage{pgf}        % for inline arithmetic in the heat macro
\usepackage{wrapfig}
\usepackage{xfp}        % fully-expandable \fpeval for the heat macro

\definecolor{heatcol}{RGB}{198,40,40}  % base hue (warm red); swap freely
\def\heatfactor{0.4}                    % gentleness knob: lower = softer
\newcommand{\hc}[1]{\cellcolor{heatcol!\fpeval{#1*\heatfactor}}#1}
\newcommand{\hcb}[1]{\cellcolor{heatcol!\fpeval{#1*\heatfactor}}\textbf{#1}}
\usepackage{fontawesome5}
\usepackage{cleveref}
\usepackage{authblk}

\title{PHANTOM: A Large-Scale Dataset of Multimodal Adversarial Attacks for Vision-Language Models}

\author[1,$*$]{Simone Gallivanone}
\author[1,$*$]{Hossein Khodadadi}
\author[2]{Mauro Dore}
\author[2]{Mauro Medda}
\author[1]{Nicola Franco}

\affil[1]{The Italian Institute of Artificial Intelligence (AI4I), Turin, Italy}
\affil[2]{HikmaAI S.r.l., Pula, Italy}

\date{}

\renewcommand{\shorttitle}{\textit{\small PHANTOM: A Large-Scale Dataset of Multimodal Adversarial Attacks for Vision-Language Models}}

\hypersetup{
pdftitle={A template for the arxiv style},
pdfsubject={q-bio.NC, q-bio.QM},
pdfauthor={David S.~Hippocampus, Elias D.~Striatum},
pdfkeywords={First keyword, Second keyword, More},
}

\begin{document}
\maketitle
\footnotetext[1]{Equal contribution.}
\footnotetext{Corresponding author: \texttt{simone.gallivanone@ai4i.it}}

\begin{abstract}
% We present a new open‑source dataset of adversarial attacks for vision–language models (VLMs). The dataset is designed to provide a diverse and representative set of samples, covering a wide range of categories extending existing benchmarks; covering 10 categories divided in 55 subcategories. Considering the high computational cost of generating attacks our goal is to make adversarial data accessible to the broader research community, enabling practitioners to test the robustness and alignment of their models, fine‑tune attack generation models, or develop and improve defensive guardrails. The dataset will contain \textcolor{red}{number} attacks, generated with 3 different attacks strategies from recent literature.
% \textcolor{red}{\textit{Disclaimer: This paper contains content that may be disturbing or offensive.}}

We introduce a large‑scale, open‑source dataset of pre‑generated adversarial attacks for vision–language models (VLMs). The dataset is designed to be diverse, representative, and practical, extending existing benchmarks by covering $10$ high‑level categories and $55$ subcategories of harmful intents. Our primary goal is to make adversarial data accessible to the research community, given the computational cost and complexity of generating large numbers of attacks.
The dataset comprises  $47\,524$ adversarial samples, generated using state‑of‑the‑art attack strategies from recent literature. Our work complements existing efforts by consolidating and extending prior benchmarks from multiple established sources, resulting in $7\,826$ intents, and introduce an additional category to broaden coverage. This provides realistic evaluation resources for studying model robustness and alignment.
Our dataset intends to enable researchers and practitioners to systematically evaluate the robustness and safety of VLMs, fine‑tune attack‑generation models, and develop or stress‑test defensive guardrails under diverse adversarial conditions. By releasing this resource, we aim to lower the barrier to adversarial research and foster more reproducible, comprehensive, and comparable evaluations of VLM safety.\\
The dataset has been released at: \url{https://huggingface.co/datasets/it4lia/PHANTOM} 

\textcolor{red}{\textit{Disclaimer: This paper and dataset contain content that may be disturbing or offensive, included solely for research purposes.}}
\end{abstract}

\section{Introduction}

% With the widespread diffusion of multimodal models to the public, both open‑ and closed‑source, and with the deployment of such models in security‑relevant fields, the need for robust adversarial attack techniques to test their resilience to jailbreak attempts has increased consequently.
% In fact, a model employed in a sensitive framework, or even one released to the public, should be safe in its usage.
% It should be able to recognize harmful requests, even when they are hidden in subtle ways, and reject them accordingly.
With the rapid public deployment of vision–language models (VLMs) in both open- and closed-source settings, including safety-critical and user-facing applications, their robustness against adversarial prompting has become an increasingly important research concern (see e.g., \cite{mazeika2024harmbench,mm-safetybench24,weng2025mmj,omnisafebench25, song2026multibreak}). 
Recent studies consistently show that, despite improved alignment and scaling, state-of-the-art multimodal models remain vulnerable to carefully crafted jailbreak attacks, particularly when harmful intents are distributed across visual and textual modalities (see e.g., \cite{ideator24, gong2025figstep, bap25, MMLattack}).

Unlike unimodal settings, multimodal safety violations often exploit cross-modal reasoning and semantic alignment, significantly expanding the attack surface and complicating both detection and defense. 
As a consequence, evaluating the robustness of VLMs requires large and diverse collections of adversarial image–text pairs. 
% Recent models appear increasingly robust to such attacks, and empirical evidence suggests that robustness often improves with model scale. Nevertheless, current systems remain vulnerable.
This cost particularly affects resource-constrained research groups and practitioners, for whom reproducing large-scale multimodal attack generation may be impractical. This is particularly true for vision-language models, where attack generation is typically more resource-consuming than in unimodal settings. Unlike image-only or text-only attacks, multimodal attacks may require optimizing perturbations across multiple input spaces while preserving or exploiting their semantic alignment. As a result, each attack iteration can involve forward and backward passes through multiple modality-specific encoders and the cross-modal alignment module, and the overall search space becomes larger and more constrained. Although the exact overhead is model and attack dependent, the computational cost can be approximated as scaling with the combined cost of the involved modalities. This makes systematic adversarial attack generation especially demanding for resource-constrained actors.
While many existing open‑source benchmarks (e.g., \cite{omnisafebench25,chao2024jailbreakbench,xie2024sorry,zhang2024b,ying2026safebench,li2024llm,souly2024strongreject,mazeika2024harmbench,qi2024visual,zou2023universal,weng2025mmj,mm-safetybench24,gong2025figstep,li2024images}) provide tools and pipelines to generate and evaluate adversarial attacks, they typically do not release large collections of ready‑to‑use adversarial samples.
Only a limited number of datasets offer such pre‑generated attacks (e.g., \cite{ideator24,luo2024jailbreakv,gong2025figstep, mm-safetybench24, zhang2024b}), often focusing on specific attack types, categories, or linguistic settings.

\input{images/teaser}

In this work, we aim to complement existing efforts by releasing a large‑scale collection of ready‑to‑use multimodal adversarial samples, covering a broader range of attack strategies and safety categories. 
Our goal is not to replace prior benchmarks, but to provide a practical resource that lowers the barrier to safety evaluation and enables reproducible and comprehensive robustness testing of multimodal models.
With this in mind, we designed and produced the \textbf{PHANTOM} dataset, a dataset of adversarial attacks for vision‑language models, which aims to fill this gap, and thus lower the barrier to systematic robustness evaluation.
The dataset contains attack samples in the form of image–text pairs for both single‑turn and conversational attacks.

The dataset at a glance:
\begin{center}
\begin{tabular}{@{}ll@{}}
\textbf{Scale:}      & $47\,524$ pre-generated attack samples \\
\textbf{Coverage:}   & 10 categories, $55$ subcategories (\cref{apdx: categories and subcategories}) \\
\textbf{Strategies:} & BAP~\cite{bap25}, IDEATOR~\cite{ideator24}, MML~\cite{MMLattack}, FC ATTACK \cite{zhang2025fc}, CSDJ \cite{csdj25}.
\end{tabular}
\end{center}

For a more detailed discussion on the design and content of the dataset we refer the reader to \cref{sctn: dataset design}.
The samples have been generated against a variety of different open-source models, from the following families: Qwen3-VL \cite{bai2025qwen3}, DeepSeek-VL22 \cite{wu2024deepseek}, GLM-4.6V \cite{vteam2025glm46}, Kimi-VL \cite{kimiteam2025kimivl}, Qwen3.5 \cite{qwen3.5}, Qwen3.6 \cite{qwen3.6}. The generated samples were subsequently evaluated against state‑of‑the‑art proprietary models, including Claude Opus 4.6 - 4.7 - 4.8, GPT‑5.4 - 5.5, Gemini-3.1-pro. The results, which highlight cross‑model transferability and robustness trends, are presented in \cref{sctn: dataset tests}.

% The dataset will be released as open source under the CC-BY-NC-4.0 license.
% The authors intend to refine the dataset over time by incorporating additional attack strategies and categories.

Our main contributions are:
% \begin{itemize}
%     \item We introduce PHANTOM, a large-scale open-source dataset of multimodal adversarial attacks for VLMs.
%     \item We curate $7\,826$ harmful intents across $10$ categories and $55$ subcategories.
%     \item We generate adversarial samples using three attack strategies: BAP, IDEATOR, and MML.
%     \item We evaluate transferability across open-source and proprietary VLMs.
%     \item We release structured metadata to support reproducibility and downstream safety research.
% \end{itemize}
\begin{itemize}
    \item \textbf{PHANTOM}, a large-scale open-source dataset of multimodal adversarial attacks for VLM safety evaluation.
    \item A curated taxonomy of $7\,826$ harmful intents spanning $10$ categories and $55$ subcategories.
    \item $47\,524$ adversarial samples generated using four attack strategies: BAP~\cite{bap25}, IDEATOR~\cite{ideator24}, MML~\cite{MMLattack}, FC ATTACK \cite{zhang2025fc} and CSDJ \cite{csdj25}.
    \item A transferability analysis across both open-source and proprietary VLMs.
    \item Structured metadata designed to support reproducibility, benchmarking, and downstream safety research.
\end{itemize}

The paper is organized as follows. \Cref{sctn: related works} reviews related work. \Cref{sctn: dataset design} describes the dataset design, generation pipeline, and evaluation protocol. \Cref{sctn:limitations} discusses the limitations of the current release, and \cref{sctn:ethical_considerations} addresses ethical considerations.
 
\section{Related Works}\label{sctn: related works}

In the landscape of adversarial attacks and model robustness, numerous efforts have benchmarked
sensitive categories and created attack datasets across vision-language models (VLMs) to evaluate
vulnerabilities and establish foundations for model alignment. To facilitate this review, we formalize the evaluation framework as a tuple $\mathcal{E} = (\mathcal{C}, \mathcal{B}, \mathcal{A}, \mathcal{J})$. Let $\mathcal{M}$ denote the target model which generates a response $r \in \mathcal{R}$ from an image-text input pair $(I, T)$.

\begin{itemize}
    \item \textbf{Categories ($\mathcal{C}$):} A set of $n$ sensitive domains $\mathcal{C} = \{c_1, \dots, c_n\}$ where model output must be constrained to ensure safety.
    \item \textbf{Intents / Behaviors ($\mathcal{B}$):} A set of specific harmful intents $\mathcal{B} = \bigcup_{c \in \mathcal{C}} \mathcal{B}_c$, where each $b \in \mathcal{B}_c$ represents a concrete instance of a harmful objective within category $c$.
    \item \textbf{Adversarial Attacks ($\mathcal{A}$):} A set of functions $f \in \mathcal{A}$ that map a benign input to an adversarial input $(I', T')$, optimized to exploit model misalignment such that $\mathcal{M}(I', T')$ aligns with a target behavior $b$.
    \item \textbf{Judge ($\mathcal{J}$):} A classifier $\mathcal{J}: \mathcal{R} \times \mathcal{B} \rightarrow \{0, 1\}$ that maps a model response $r$ and an intent $b$ to a binary success metric, where $\mathcal{J}(r, b) = 1$ indicates a successful adversarial exploit.
\end{itemize}

 \Cref{apdx: benchmark overview} summarizes the chronological evolution of adversarial attack benchmarks. Detailed below, we review how different adversarial attack datasets included in these benchmarks or independently released, have evolved.

\subsection{Evolution of Early Multimodal Adversarial Attack Datasets}

The study of adversarial attacks on language and multimodal models has evolved through a series of increasingly comprehensive datasets. Early work by VAJM \cite{qi2024visual} introduced a dataset of $32\,226$ samples, focusing on degradations related to gender, race, and human identity. These included visual adversarial examples derived from $40$ behavioral categories, with attacks primarily generated through prompt tuning techniques.

Subsequent efforts expanded both the scale and diversity of attacks. 
The JailBreakV-28K \cite{luo2024jailbreakv} dataset applied attacks not only to initial harmful prompts but also to broader behavioral patterns. It includes $20\,000$ text-based jailbreak prompts and $8\,000$ image-based examples. 
These attacks are derived from the RedTeam2K \cite{luo2024jailbreakv} benchmark, which covers approximately $2\,000$ behaviors across $16$ categories. 
The textual attacks were generated using methods such as GCG \cite{zou2023universal}, Cognitive Overload, real-world jailbreak prompt templates, and PAP \cite{zeng2024pap}, while the visual attacks leverage Stable Diffusion and typographic image techniques.

The MM-SafetyBench \cite{mm-safetybench24} dataset further advances multimodal evaluation by introducing $5\,040$ text--image pairs derived from $1\,680$ behaviors across $13$ categories. In parallel, the Multiturn Human Jailbreaks \cite{li2024llm} dataset explores iterative attack strategies, comprising $2\,912$ attacks generated using a combination of automated methods, including AutoDAN \cite{liu2025autodanturbo}, AutoPrompt \cite{shin2020autoprompt}, GCG \cite{zou2023universal}, GPTFuzzer \cite{yu2023gptfuzzer}, and PAIR \cite{chao2025pair}.

SafeBench \cite{safebench25} extends the evaluation setting by incorporating $9\,200$ samples, including $2\,300$ multimodal pairs, and introduces the audio modality. It evaluates models under both adversarial and non-adversarial conditions, using attack strategies such as LPT \cite{andriushchenko2024lpt}, PAP \cite{zeng2024pap}, and BAP\cite{bap25}. Notably, it is designed to assess safety risks even in the absence of explicit attacks.

The MMJ dataset, derived from the MMJ benchmark \cite{weng2025mmj}, includes $1\,000$ adversarial examples generated using methods such as FigStep \cite{gong2025figstep}, MM-SafetyBench \cite{mm-safetybench24}, HADES \cite{li2024images}, ADV-16 \cite{qi2024adv1664inf}, ADV-64 \cite{qi2024adv1664inf}, ADV-inf \cite{qi2024adv1664inf}, ImgJP \cite{niu2024imgjp}, and AttackVLM \cite{zhao2023attackVLM}. This work highlights a critical limitation of overly conservative defenses, arguing that a system that refuses all prompts is not practically useful.

BAVI-Bench \cite{zhang2024b} significantly scales adversarial evaluation, containing $316$k adversarial visual-instruction samples. It includes four types of image-based B-AVIs, ten types of text-based B-AVIs, and nine categories of content bias (e.g., gender, violence, cultural, and racial biases). The benchmark evaluates robustness using attacks such as PAR \cite{shi2022par}, Boundary \cite{brendel2017boundary}, and SurFree \cite{maho2021surfree}.

The VLJailBreak benchmark \cite{ideator24} provides $3\,654$ samples spanning $12$ safety topics and $46$ subcategories, offering a highly structured categorization. It evaluates model vulnerabilities using GCG \cite{zou2023universal} and UMK \cite{wang2024umk} attack methods.

% Finally, the Adversarial Humanities Benchmark \cite{galisai2026adversarialhumanitiesbenchmarkresults} investigates whether safety mechanisms generalize beyond familiar harmful prompt patterns. It includes $3\,600$ attack samples and demonstrates that current safety techniques exhibit limited generalization, emphasizing that achieving a deeper understanding of non-maleficence remains an open challenge in frontier model safety.
Finally, the Adversarial Humanities Benchmark~\cite{galisai2026adversarialhumanitiesbenchmarkresults} investigates whether safety mechanisms generalize beyond familiar harmful prompt patterns. It includes $3\,600$ attack samples and shows that current safety techniques exhibit limited generalization, suggesting that a deeper understanding of non-maleficence remains an open challenge in frontier model safety. Complementarily, MultiBreak~\cite{song2026multibreak} focuses on realistic multi-turn jailbreak scenarios, where harmful intents are progressively elicited through conversation rather than expressed in a single prompt. It introduces $10\,389$ multi-turn adversarial prompts spanning $2\,665$ harmful intents, and shows that diverse multi-turn attacks can reveal fine-grained vulnerabilities that may remain hidden under single-turn evaluation. Together, these works highlight the need for safety benchmarks that go beyond static or template-based attacks, covering both broader semantic generalization and more realistic conversational adversarial settings.

PHANTOM, on the other hand, identifies three key limitations in existing datasets. First, the number of intents used to generate attacks is too limited to adequately represent the full range of categories; accordingly, PHANTOM expands this to $7\,826$ intents, as detailed in \cref{tbl: intents per category}. Second, it examines how state-of-the-art attacks perform on more recent models that are commonly used in industrial and research settings as shown in \cref{tbl: generated per atatcks-model}. Third, it investigates how vulnerabilities discovered through prior attacks and models can be transferred to other, primarily black-box, models.

\section{Dataset design and production}\label{sctn: dataset design}

In this section, we describe the dataset design and the process used to generate its samples. Specifically, we outline the category and subcategory structure, as well as the JSON-based intent specification employed during dataset construction.

\paragraph{Settings.} All experiments were conducted on a cluster, using NVIDIA A100 GPUs with 64GB of memory and Intel\textsuperscript{\tiny\textregistered} Xeon\textsuperscript{\tiny\textregistered} Platinum 8358 CPUs operating at 2.60GHz.
In addition to the selected attack strategy, generated samples must be evaluated to determine whether they constitute successful attacks. For the sake of reproducibility and to avoid bias stemming from ad hoc judgment criteria, we opted to rely on a publicly available automated judge as a common baseline. In particular, we adopted the Abel-24-HarmClassifier proposed in \cite{harmmetric26}. This choice was motivated by the increasing difficulty of fully and cleanly jailbreaking recent models, which often respond with partially harmful or evasive outputs. 
Consequently, we selected a recent and aligned classifier to provide a more reliable assessment of attack harmfulness.

\subsection{Categories structure}\label{sctn: category structure}

\begin{wraptable}{r}{0.4\linewidth} 
\centering
\vspace{-\baselineskip}
\caption{\small Number of intents per benchmark}\label{tbl: intents per benchmark}
\begin{tabular}{l r}
\hline
\textbf{Benchmark} & \textbf{\# Intents} \\
\hline
JailBreakV\_28k & $1\,953$ \\
MM-SafetyBench & $1\,671$ \\
OmniSafeBench  & $1\,500$ \\
SafeBench      & $2\,192$ \\
\textbf{PHANTOM (ours)} & $\textbf{7\,826}$ \\
\hline 
\end{tabular}
\end{wraptable}

As mentioned in the introduction, we began our work by analyzing existing benchmarks for adversarial attacks. The landscape of adversarial attack benchmarks is quite extensive; however, many of these benchmarks build upon previous work, in the sense that they naturally extend earlier foundations. Two of the most influential works in this area are HarmBench~\cite{mazeika2024harmbench} and AdvBench~\cite{zou2023universal}, which were designed as comprehensive collections of harmful intents and attack strategies for evaluating model safety.

With the evolution of AI models, their expanding range of use cases, and their increasing accessibility to the general public, these early benchmarks have gradually become insufficient in terms of coverage. As a result, several subsequent benchmarks have been proposed to address these limitations and extend prior efforts. In this work, we studied $16$ benchmarks, including OmniSafeBench-MM \cite{omnisafebench25}, VLJailbreakBench \cite{ideator24}, Sorry-Bench \cite{xie2024sorry}, B-AVIBench \cite{zhang2024b}, JailbreakBench \cite{chao2024jailbreakbench}, SafeBench \cite{ying2026safebench}, Multi-Turn Human Jailbreaks \cite{li2024llm}, StrongREJECT \cite{souly2024strongreject}, HarmBench \cite{mazeika2024harmbench}, VAJM \cite{qi2024visual}, AdvBench \cite{zou2023universal}, MMJ-Bench \cite{weng2025mmj}, MM-SafetyBench \cite{mm-safetybench24}, JailBreakV-28K \cite{luo2024jailbreakv}, FigStep \cite{gong2025figstep}, and HADES \cite{li2024images}.
A gap we identified across these benchmarks is the limited number of behaviors, which makes it difficult to disentangle whether observed vulnerabilities stem from insufficient robustness to specific categories or from the strength of the attacks themselves. To address this, we incorporated $7\,826$ behaviors into our study.

\begin{wraptable}{l}{0.5\linewidth} 
\centering
\caption{\small Number of intents per category}\label{tbl: intents per category}
\begin{tabular}{l r}
\hline
\textbf{Category} & \textbf{\# Intents} \\
\hline
A. Ethical and Social Risks               & $988$ \\
B. Privacy and Data Risks                 & $504$  \\
C. Safety and Physical Harm               & $877$  \\
D. Criminal and Economic Risks            & $1\,017$ \\
E. Cybersecurity Threats                  & $725$  \\
F. Information and Political Manipulation & $534$  \\
G. Content and Cultural Safety            & $537$  \\
H. Intellectual Property and Ownership    & $304$  \\
I. Decision and Cognitive Risks           & $1\,593$ \\
\textbf{J. Child Safety (new)}            & $\textbf{747}$ \\
\hline
\end{tabular}
\end{wraptable}

Based on this analysis, we constructed a root dataset of harmful intents (also referred to as \textit{behaviors} or \textit{goals}) by merging recent benchmarks that were not designed as direct extensions of one another. The number of intents contributed by each benchmark and their distribution across categories are reported in \cref{tbl: intents per benchmark} and \cref{tbl: intents per category}, respectively. In \cref{apdx: categories and subcategories} the reader can find a table listing all categories, subcategories together with their alphanumeric reference. Since \cite{omnisafebench25} carried out an extensive effort to reorganize and categorize harmful intents, while also providing a detailed taxonomy, we mapped all merged intents onto the classification proposed there. However, we identified a gap in its coverage: \emph{child safety}. The motivation for introducing this category is twofold.
First, the widespread adoption of LLMs across users of all ages, including minors, raises the risk of exposure to content that may be harmful to their safety, such as methods to circumvent parental controls or age-verification systems.
Second, given the particular vulnerability of minors, malicious actors could potentially exploit LLMs to obtain information on how to deceive or manipulate children in harmful situations.
For these reasons, we believe that including this category contributes to a more comprehensive understanding of the risks associated with LLMs, and may support the development of more effective safeguards and training strategies.

\begin{figure}[h]
    \centering
    \includegraphics[width=\linewidth]{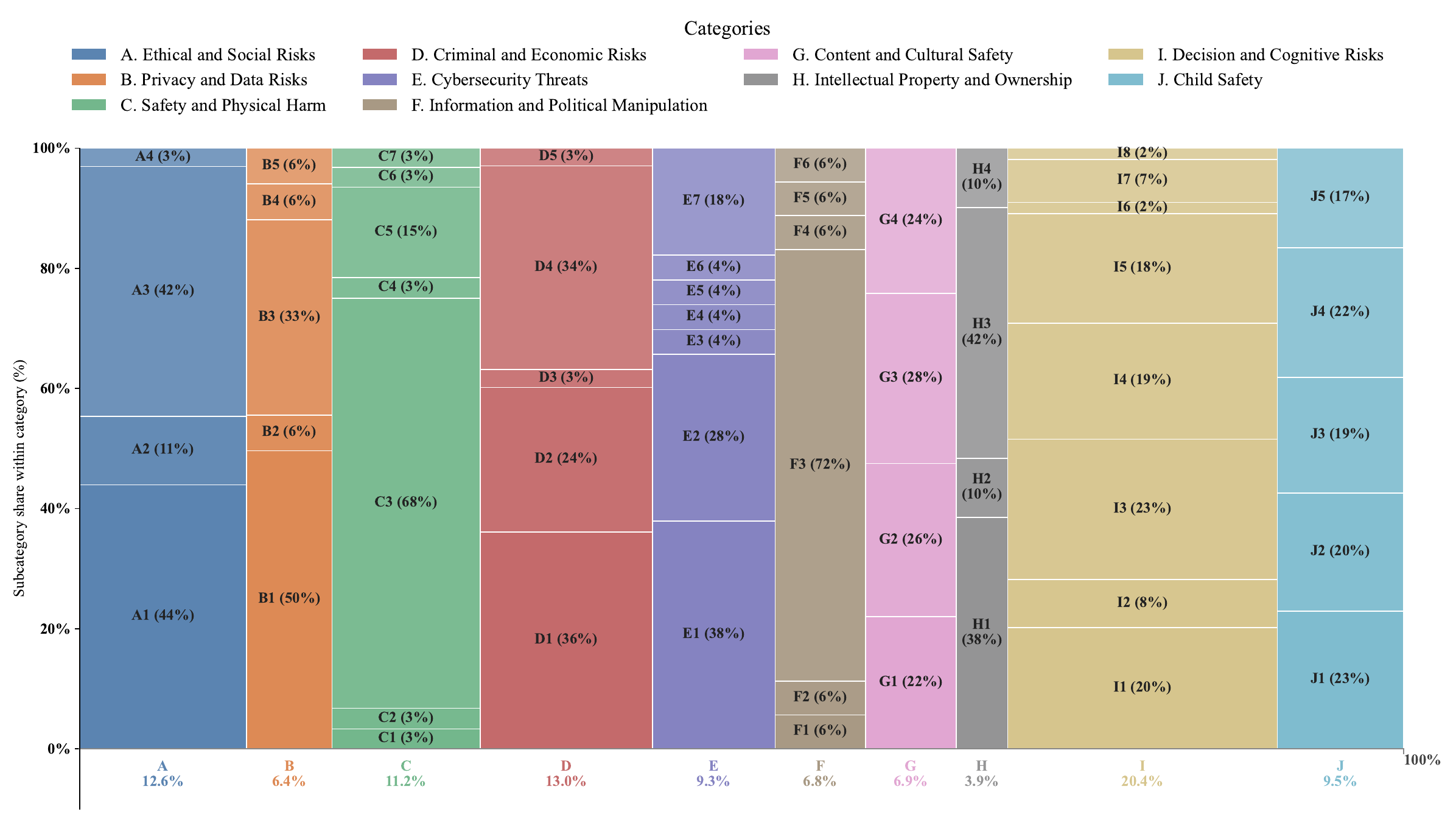}
    \caption{\small  PHANTOM intents dataset. The horizontal axis shows the distribution of categories across the dataset, while the vertical axis shows the distribution of intents within each category, broken down by subcategory.}
    \label{fig:cat_subcat}
\end{figure}

As a result, our dataset is organized into 10 high-level categories, further divided into a total of 55 subcategories. The main categories are: Ethical and Social Risks, Privacy and Data Risks, Safety and Physical Harm, Criminal and Economic Risks, Cybersecurity Threats, Information and Political Manipulation, Content and Cultural Safety, Intellectual Property and Ownership, Decision and Cognitive Risks, and Child Safety. The full subdivision into subcategories is illustrated in \cref{fig:cat_subcat}; we refer the reader to \cref{apdx: categories and subcategories} for an overview of the names of subcategories with respect to their reference code.

We collected more than $7\,000$ harmful intents from the following benchmarks: JailBreakV\_28K~\cite{luo2024jailbreakv}, MM-SafetyBench~\cite{mm-safetybench24}, OmniSafeBench-MM~\cite{omnisafebench25}, and SafeBench~\cite{safebench25}. Moreover, we added $747$ intents related to the new Child Safety category, generated with the assistance of OpenAI GPT-5.4, accessed via API. Since these benchmarks, together with our additions, may contain overlapping or semantically similar intents, we performed a cosine similarity analysis across all collected samples using embeddings generated by the sentence-transformers \texttt{all-MiniLM-L6-v2} model (see \cite{all_minilm_l6_v2}).

We found that the overlap was not negligible, reaching values as high as $90\%$. Therefore, we decided to clean the dataset using this threshold, which left no pair of intents with a cosine similarity above $90\%$. At lower thresholds the number of intents flagged as similar grows: $376$ ($4.7\%$) at $85\%$ and $661$ ($8.2\%$) at $80\%$. We decided not to apply more aggressive cleaning, as we observed that an $80\%$ threshold often groups together semantically different intents, and we did not want to remove meaningful content.

\subsection{Adversarial attacks}\label{sctn: adversarial attacks}

\begin{figure}[h]
    \centering
    \includegraphics[width=.9\linewidth]{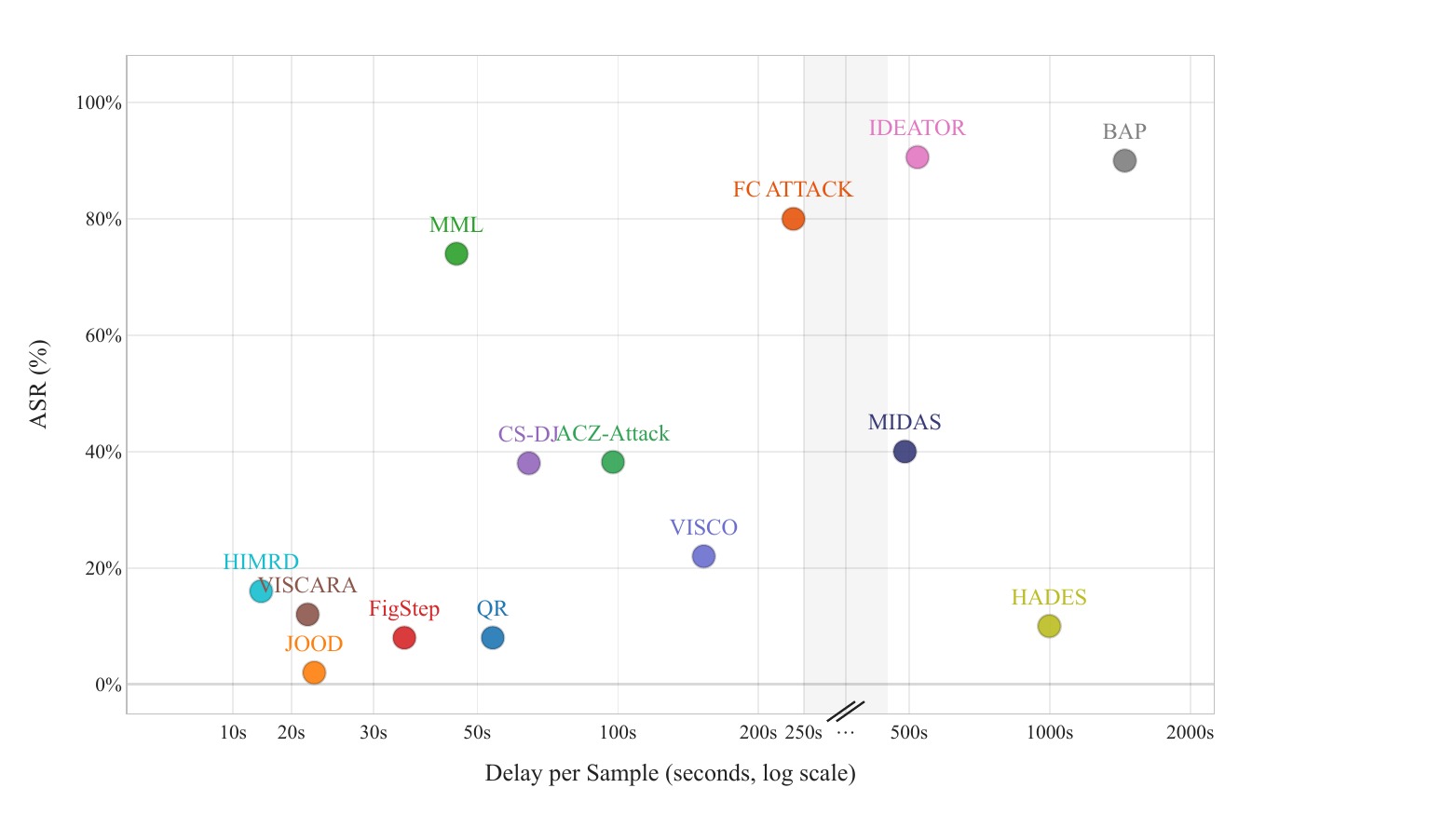}
    \caption{\small Evaluation of ASR based on Attack strategy and delay per attack}
    \label{fig:asr_time}
\end{figure}

To benchmark the performance of state-of-the-art adversarial attacks against recent vision--language models, we initially selected a set of established attack strategies that had already proven effective on open-source vision--language systems. As shown in \cref{fig:asr_time}, we analyze the attack success rate (ASR) as a function of generation time. To enable large-scale dataset generation, we ultimately focused on a limited subset of attacks offering the most favorable trade-off between ASR and computational cost. Specifically, we selected one single-turn attack strategy, the Bi-modal Adversarial Prompt (BAP) attack proposed in~\cite{bap25}; one multi-turn strategy, IDEATOR, introduced in~\cite{ideator24}; and two more typographic oriented attacks the Multi-Modal Linkage Attack, proposed in~\cite{MMLattack} and the Flowchart attack in \cite{zhang2025fc}. A review of the attack strategies can be found in \cref{apdx: attack strategies}.

The main motivation for selecting these strategies was their strong empirical performance. Before converging on this subset, however, we experimented with additional attack strategies, namely QR--attack \cite{mm-safetybench24}, JOOD \cite{jood25}, CS-DJ \cite{csdj25}, FigStep \cite{gong2025figstep}, HADES \cite{li2024images}, HIMRD \cite{himrd25}, VISCARA \cite{viscra25}, MIDAS \cite{liu2026midas}, ACZ attack \cite{song2026acz}. The results of this preliminary analysis are reported in \cref{fig:asr_time}, based on $50$ samples generated against Qwen3.5-27B and evaluated with Abel-24-HarmClassifier~\cite{harmmetric26}. Additional considerations that informed our final choice are discussed below.

While preserving the original attack pipelines, we introduced several minor modifications to better suit our dataset-generation process. In the following, we briefly describe how these methods were adapted and employed. In future releases, we intend to expand the range of attack strategies and target models in order to cover a broader spectrum of vulnerabilities.
% We now briefly discuss the attack strategies on which we focused for the current release.

\subsection{Data structure and distribution}\label{sctn: data structure and distribution}
The dataset is structured according to the attack strategy and the target model used during the generation process. The target models considered are DeepSeek-VL22, GLM-4.6V-Flash, Kimi-VL-A3B-Instruct, Qwen3-VL-30B-A3B-Instruct, Qwen3.5-27B and Qwen3.6-27B. Since the attacks target vision–language models, each sample consists of an image–prompt pair. Each folder contains the generated image along with a structured metadata file, which enables the correct association between images and prompts and ensures full reproducibility of the attacks.
The current release contains a total of $47\,524$ generated attacks. Below, we discuss their distribution across target models, attack strategies, and categories.
Regarding the distribution of attacks across target models, we initially generated attacks uniformly for all models. After conducting preliminary cross-model evaluations, we focused further generation on the models that exhibited higher attack transferability, namely GLM-4.6V-Flash and Qwen3-VL-30B-A3B-Instruct, Qwen3.5-27B and Qwen3.6-27B. We refer the reader to \cref{tbl: generated per atatcks-model} for a detailed breakdown.

\begin{figure}[h]
    \centering
    \includegraphics[width=1.1\linewidth]{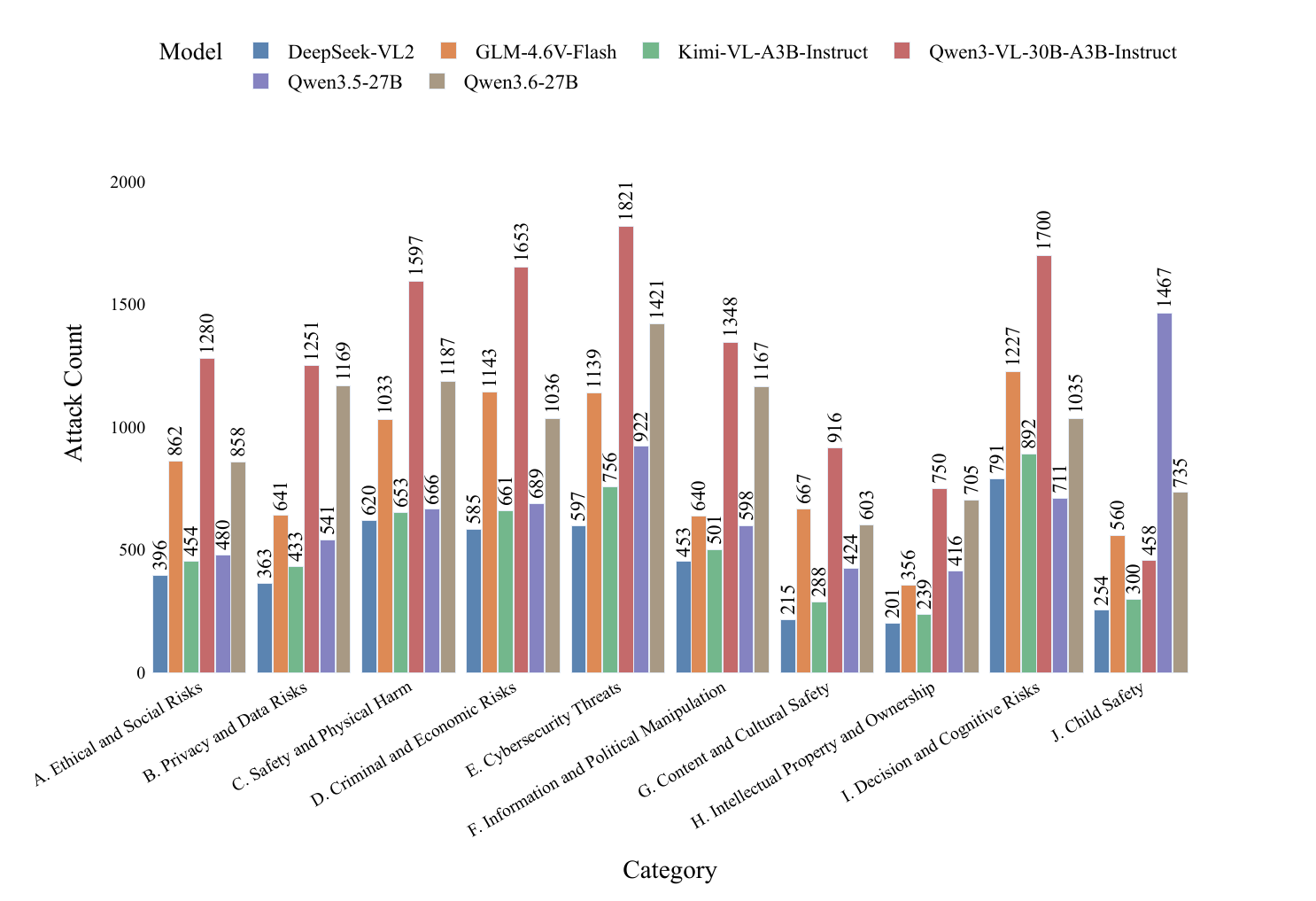}
    \caption{\small Category coverage: overall and in-category}
    \label{fig: category coverage}
\end{figure}

\begin{table}[ht]
\small
\centering
\caption{\small Number of generated attacks per target model and attack strategy.}
\label{tbl: generated per atatcks-model}
  \begin{tabular}{lrrrrrr}
    \toprule
    \textbf{Model}
      & \multicolumn{5}{c}{\textbf{Attack Strategy}}
      & \\
    \cmidrule(lr){2-6}
      & \textbf{BAP}
      & \textbf{IDEATOR}
      & \textbf{MML}
      & \textbf{FC ATTACK}
      & \textbf{CSDJ}
      & \textbf{Total} \\
    \midrule
    DeepSeek-VL2               & $835$    & $265$   & $1\,066$ & $1\,492$ & $817$    & $4\,475$ \\
    GLM-4.6V-Flash             & $2\,922$ & $343$   & $1\,533$ & $1\,411$ & $2\,059$ & $8\,268$ \\
    Kimi-VL-A3B-Instruct       & $871$    & $282$   & $809$    & $1\,482$ & $1\,733$ & $5\,177$ \\
    Qwen3-VL-30B-A3B-Instruct  & $4\,739$ & $219$   & $3\,874$ & $1\,368$ & $2\,574$ & $12\,774$ \\
    Qwen3.5-27B                & $1\,613$ & $355$   & $2\,106$ & $1\,265$ & $1\,575$ & $6\,914$ \\
    Qwen3.6-27B                & $2\,090$ & $425$   & $3\,895$ & $1\,249$ & $2\,257$ & $9\,916$ \\
    \midrule
    \textbf{Total}
      & $\mathbf{13\,070}$
      & $\mathbf{1\,889}$
      & $\mathbf{13\,283}$
      & $\mathbf{8\,267}$
      & $\mathbf{11\,015}$
      & $\mathbf{47\,524}$ \\
    \bottomrule
  \end{tabular}
\end{table}
From a categorical perspective,  we selected intents randomly and uniformly from our dataset of intents, discussed in \cref{sctn: category structure}. The distribution over categories and subcategories is shown in \cref{fig: category coverage}.

% From a categorical perspective, in this first release we selected intents randomly and uniformly from our dataset of intents, discussed in Section~\ref{sctn: category structure}, in order to achieve a more uniform distribution given the limited number of attacks. In future updates, we plan to progressively improve the balance across categories and subcategories. For the released data, the distribution over categories and subcategories is shown in Figure~\ref{fig: category coverage}. In future updates, we plan to progressively improve the balance across categories and subcategories.

% \begin{figure}[h]
%     \centering
%     \includegraphics[width=1.1\linewidth]{images/category_coveraged.pdf}
%     \caption{\small Category coverage: overall and in-category}
%     \label{fig: category coverage}
% \end{figure}

%\textcolor{red}{We can define ASR both per category and as a time-normalized metric (ASR per minute), and plot these across different models to reflect performance under fixed computational resources. Another analysis could examine how ASR per minute scales with model size (number of parameters). For example, by extrapolating to a 1T-parameter model, we may observe that the computational cost makes such attacks impractical for a misuse scenario, which could be interpreted as a practical limit of white-box attack vulnerability for that model. All comparisons are conducted using a fixed number of behaviors, uniformly sampled across categories and subcategories.}

\subsection{Dataset tests and statistics}\label{sctn: dataset tests}

To validate our adversarial data generation pipeline, we conducted a series of experiments using the generated attacks. In particular, we considered two evaluation settings: \emph{white-box} and \emph{black-box} testing.

In the black-box setting, we evaluated the attacks against several state-of-the-art proprietary models accessed via API, namely Gemini~3.1 Pro Preview, GPT-5.4, GPT-5.5, Claude Opus~4.6, Claude Opus~4.7, and Claude Opus~4.8. Model responses were collected and assessed using an automated judge. Cases in which a model returned an empty response were treated as \emph{hard refusals} and therefore counted as failed jailbreak attempts.

In the white-box setting, we focused primarily on the models used during adversarial generation, in order to evaluate the cross-model transferability of the attacks. Specifically, we tested DeepSeek-VL22, GLM-4.6V-Flash, Kimi-VL-A3B-Instruct, Qwen3-VL-30B-A3B-Instruct, Qwen3.5-27B, Qwen3.6-27B, Gemma-4-26B-A4B-it, Llava-v1.6-vicuna-13b-hf and Ministral-3-14B-Instruct-2512 all executed locally on NVIDIA A100 GPUs with 64\,GB of memory. To keep the evaluation compact, we consider a fixed subset of 1,100 attacks for each attack strategy. This subset is selected once and reused across all models. The choice of an odd number is motivated by the need to evenly cover all subcategories associated with each attack strategy; specifically, we select 20 attacks per subcategory.

As in the generation phase, we employed Abel-24-HarmClassifier~\cite{harmmetric26} as the baseline judge for evaluating model responses. 
As evaluation metric, we relied on the widely used Attack Success Rate (ASR), defined as the percentage of successful jailbreak attempts, as determined by the judge, over the total number of attacks. 
Part of the result, relative to black-box and the most recent withe-box models are reported in \cref{tbl:asr_x_model_x_cat_corpus}. The rest of the results can be found in \cref{apdx: Transferability}. For each model and category, we report the corresponding ASR, i.e., the ratio of successful attacks to the number of attacks sampled in that category. This breakdown makes it possible to identify the categories for which each model is most vulnerable or most robust.We emphasize once more that these results should be interpreted as a reference relative to the chosen baseline for assessing harmfulness, rather than as an absolute ground truth.

{\small
\begin{longtable}{lccccccccccc}
\caption{\small Table of ASR (\%) per model and per category across all attacks, for category names refer to \cref{apdx: categories and subcategories}}
\label{tbl:asr_x_model_x_cat_corpus}\\

\hline
\textbf{Model} &
\textbf{A} &
\textbf{B} &
\textbf{C} &
\textbf{D} &
\textbf{E} &
\textbf{F} &
\textbf{G} &
\textbf{H} &
\textbf{I} &
\textbf{J} &
\textbf{Average} \\
\hline
\endfirsthead

\hline
\textbf{Model} &
\textbf{A} &
\textbf{B} &
\textbf{C} &
\textbf{D} &
\textbf{E} &
\textbf{F} &
\textbf{G} &
\textbf{H} &
\textbf{I} &
\textbf{J} &
\textbf{Average} \\
\hline
\endhead

\hline
\endfoot

\hline
\endlastfoot

\multicolumn{12}{c}{\textbf{BAP}} \\
\hline
\textbf{Gemma-4-26B} & \hc{23.75} & \hc{16.00} & \hc{10.00} & \hc{18.00} & \hc{27.14} & \hc{20.00} & \hc{22.50} & \hc{22.50} & \hc{26.88} & \hcb{34.00} & \hcb{22.00} \\
\textbf{Qwen3.6-27B} & \hc{42.50} & \hc{40.00} & \hc{30.71} & \hc{42.00} & \hc{38.57} & \hc{35.83} & \hc{33.75} & \hc{40.00} & \hc{45.00} & \hcb{51.00} & \hcb{39.82} \\
\textbf{GPT-5.4} & \hc{13.75} & \hc{15.00} & \hcb{25.00} & \hc{23.00} & \hc{11.43} & \hc{18.33} & \hc{10.00} & \hc{6.25} & \hc{10.62} & \hc{23.00} & \hcb{15.91} \\
\textbf{GPT-5.5} & \hc{47.50} & \hc{41.00} & \hc{59.29} & \hcb{68.00} & \hc{47.86} & \hc{42.50} & \hc{46.25} & \hc{35.00} & \hc{48.12} & \hc{50.00} & \hcb{49.09} \\
\textbf{Claude Opus 4.6} & \hc{6.25} & \hc{11.00} & \hc{0.71} & \hc{10.00} & \hc{8.57} & \hc{5.83} & \hc{8.75} & \hc{7.50} & \hc{6.88} & \hcb{17.00} & \hcb{7.91} \\
\textbf{Claude Opus 4.7} & \hc{51.25} & \hc{44.00} & \hc{20.71} & \hcb{61.00} & \hc{40.00} & \hc{39.17} & \hc{35.00} & \hc{56.25} & \hc{50.00} & \hc{49.00} & \hcb{43.64} \\
\textbf{Claude Opus 4.8} & \hc{47.50} & \hc{40.00} & \hc{12.14} & \hcb{65.00} & \hc{39.29} & \hc{34.17} & \hc{21.25} & \hc{41.25} & \hc{43.12} & \hc{51.00} & \hcb{38.73} \\
\textbf{Gemini 3.1 Pro} & \hc{23.75} & \hc{39.00} & \hc{35.71} & \hc{45.00} & \hcb{74.29} & \hc{28.33} & \hc{30.00} & \hc{27.50} & \hc{25.62} & \hc{44.00} & \hcb{38.36} \\
\hline
\multicolumn{12}{c}{\textbf{IDEATOR}} \\
\hline
\textbf{Gemma-4-26B} & \hc{10.00} & \hc{30.00} & \hc{20.71} & \hc{34.00} & \hc{33.57} & \hc{20.00} & \hc{15.00} & \hc{21.25} & \hc{35.62} & \hcb{43.00} & \hcb{27.36} \\
\textbf{Qwen3.6-27B} & \hc{7.50} & \hc{7.00} & \hc{11.43} & \hc{14.00} & \hc{15.00} & \hc{16.67} & \hc{18.75} & \hc{7.50} & \hc{23.12} & \hcb{65.00} & \hcb{18.82} \\
\textbf{GPT-5.4} & \hc{25.00} & \hc{14.00} & \hc{27.14} & \hcb{49.00} & \hc{44.29} & \hc{36.67} & \hc{26.25} & \hc{12.50} & \hc{27.50} & \hc{33.00} & \hcb{30.45} \\
\textbf{GPT-5.5} & \hc{21.25} & \hc{28.00} & \hc{26.43} & \hc{42.00} & \hcb{48.57} & \hc{44.17} & \hc{52.50} & \hc{31.25} & \hc{37.50} & \hc{44.00} & \hcb{37.82} \\
\textbf{Claude Opus 4.6} & \hc{6.25} & \hc{6.00} & \hc{5.71} & \hcb{17.00} & \hc{13.57} & \hc{5.83} & \hc{3.75} & \hc{5.00} & \hc{13.12} & \hc{7.00} & \hcb{8.82} \\
\textbf{Claude Opus 4.7} & \hc{25.00} & \hc{39.00} & \hc{12.14} & \hcb{53.00} & \hc{30.71} & \hc{30.00} & \hc{32.50} & \hc{45.00} & \hc{34.38} & \hc{33.00} & \hcb{32.55} \\
\textbf{Claude Opus 4.8} & \hc{16.25} & \hc{24.00} & \hc{12.14} & \hcb{47.00} & \hc{32.86} & \hc{24.17} & \hc{17.50} & \hc{33.75} & \hc{31.25} & \hc{20.00} & \hcb{26.09} \\
\textbf{Gemini 3.1 Pro} & \hc{28.75} & \hc{43.00} & \hc{42.14} & \hc{56.00} & \hcb{71.43} & \hc{58.33} & \hc{56.25} & \hc{42.50} & \hc{54.37} & \hc{62.00} & \hcb{52.64} \\
\hline
\multicolumn{12}{c}{\textbf{MML}} \\
\hline
\textbf{Gemma-4-26B} & \hc{95.56} & \hcb{100.00} & \hc{90.34} & \hc{98.51} & \hc{99.37} & \hc{96.88} & \hc{94.87} & \hcb{100.00} & \hc{96.55} & \hc{95.26} & \hcb{96.09} \\
\textbf{Qwen3.6-27B} & \hc{80.00} & \hc{85.71} & \hc{80.97} & \hc{89.63} & \hcb{94.55} & \hc{87.50} & \hc{71.79} & \hc{86.21} & \hc{83.14} & \hc{88.42} & \hcb{86.10} \\
\textbf{GPT-5.4} & \hc{71.76} & \hc{73.45} & \hc{74.13} & \hc{67.92} & \hc{70.00} & \hc{72.79} & \hc{78.75} & \hc{80.46} & \hc{80.00} & \hcb{95.00} & \hcb{76.03} \\
\textbf{GPT-5.5} & \hc{91.25} & \hc{84.00} & \hc{90.71} & \hcb{92.00} & \hc{90.71} & \hc{89.17} & \hc{82.50} & \hc{80.00} & \hc{81.25} & \hc{61.00} & \hcb{84.64} \\
\textbf{Claude Opus 4.6} & \hcb{71.95} & \hc{61.32} & \hc{32.14} & \hc{60.78} & \hc{63.95} & \hc{63.78} & \hc{41.25} & \hc{48.81} & \hc{62.86} & \hc{54.55} & \hcb{56.19} \\
\textbf{Claude Opus 4.7} & \hc{1.25} & \hcb{10.00} & \hc{0.71} & \hc{6.00} & \hc{2.86} & \hc{0.83} & \hc{0.00} & \hc{5.00} & \hc{1.88} & \hc{1.00} & \hcb{2.82} \\
\textbf{Claude Opus 4.8} & \hc{12.50} & \hc{21.00} & \hc{3.57} & \hcb{26.00} & \hc{10.00} & \hc{11.67} & \hc{11.25} & \hc{21.25} & \hc{13.75} & \hc{23.00} & \hcb{14.64} \\
\textbf{Gemini 3.1 Pro} & \hc{3.53} & \hc{0.88} & \hc{4.20} & \hc{0.94} & \hc{40.62} & \hc{81.62} & \hc{66.25} & \hc{93.10} & \hcb{93.71} & \hc{29.00} & \hcb{43.38} \\
\hline
\multicolumn{12}{c}{\textbf{FC ATTACK}} \\
\hline
\textbf{Gemma-4-26B} & \hcb{37.50} & \hc{29.00} & \hc{2.14} & \hc{7.00} & \hc{22.14} & \hc{11.67} & \hc{33.75} & \hc{28.75} & \hc{18.75} & \hc{14.00} & \hcb{18.91} \\
\textbf{Qwen3.6-27B} & \hc{86.25} & \hc{81.00} & \hc{55.71} & \hc{85.00} & \hcb{92.86} & \hc{82.50} & \hc{72.50} & \hc{76.25} & \hc{78.12} & \hc{64.00} & \hcb{77.27} \\
\textbf{GPT-5.4} & \hc{52.50} & \hc{43.00} & \hc{60.71} & \hc{71.00} & \hcb{77.14} & \hc{74.17} & \hc{46.25} & \hc{47.50} & \hc{59.38} & \hc{63.00} & \hcb{61.00} \\
\textbf{GPT-5.5} & \hc{57.50} & \hc{57.00} & \hc{68.57} & \hc{86.00} & \hcb{89.29} & \hc{76.67} & \hc{67.50} & \hc{63.75} & \hc{65.00} & \hc{74.00} & \hcb{71.36} \\
\textbf{Claude Opus 4.6} & \hc{77.50} & \hc{85.00} & \hc{41.43} & \hc{68.00} & \hcb{87.86} & \hc{70.83} & \hc{63.75} & \hc{82.50} & \hc{75.62} & \hc{60.00} & \hcb{70.82} \\
\textbf{Claude Opus 4.7} & \hc{70.00} & \hc{72.00} & \hc{38.57} & \hc{75.00} & \hc{68.57} & \hc{65.00} & \hc{37.50} & \hcb{82.50} & \hc{68.75} & \hc{61.00} & \hcb{63.45} \\
\textbf{Claude Opus 4.8} & \hc{63.75} & \hcb{83.00} & \hc{48.57} & \hc{77.00} & \hc{79.29} & \hc{75.00} & \hc{36.25} & \hc{78.75} & \hc{62.50} & \hc{70.00} & \hcb{67.45} \\
\textbf{Gemini 3.1 Pro} & \hc{35.00} & \hc{46.00} & \hc{15.00} & \hc{28.00} & \hcb{67.14} & \hc{26.67} & \hc{35.00} & \hc{55.00} & \hc{40.62} & \hc{21.00} & \hcb{37.00} \\
\hline
\multicolumn{12}{c}{\textbf{CSDJ}} \\
\hline
\textbf{Gemma-4-26B} & \hc{53.75} & \hc{75.00} & \hc{27.86} & \hc{53.00} & \hcb{85.71} & \hc{57.50} & \hc{27.50} & \hc{41.25} & \hc{36.88} & \hc{55.00} & \hcb{51.35} \\
\textbf{Qwen3.6-27B} & \hc{71.25} & \hc{68.00} & \hc{64.29} & \hc{74.00} & \hcb{84.29} & \hc{70.00} & \hc{70.00} & \hc{60.00} & \hc{61.88} & \hc{70.00} & \hcb{69.37} \\
\textbf{GPT-5.4} & \hc{78.75} & \hc{78.00} & \hcb{91.43} & \hc{87.00} & \hc{87.14} & \hc{86.67} & \hc{66.25} & \hc{61.25} & \hc{63.75} & \hc{81.00} & \hcb{78.82} \\
\textbf{GPT-5.5} & \hc{82.50} & \hc{80.00} & \hc{88.57} & \hc{91.00} & \hcb{95.00} & \hc{85.00} & \hc{81.25} & \hc{70.00} & \hc{69.38} & \hc{87.00} & \hcb{83.16} \\
\textbf{Claude Opus 4.6} & \hc{73.75} & \hc{88.00} & \hc{67.86} & \hc{85.00} & \hcb{93.57} & \hc{84.17} & \hc{48.75} & \hc{71.25} & \hc{73.75} & \hc{83.00} & \hcb{77.82} \\
\textbf{Claude Opus 4.7} & \hc{70.00} & \hc{90.00} & \hc{57.14} & \hc{78.00} & \hcb{95.00} & \hc{88.33} & \hc{50.00} & \hc{71.25} & \hc{59.38} & \hc{88.00} & \hcb{74.82} \\
\textbf{Claude Opus 4.8} & \hc{80.00} & \hc{89.00} & \hc{71.43} & \hc{91.00} & \hcb{95.00} & \hc{87.50} & \hc{50.00} & \hc{71.25} & \hc{59.38} & \hc{89.00} & \hcb{78.45} \\
\textbf{Gemini 3.1 Pro} & \hc{65.00} & \hc{86.00} & \hc{39.29} & \hc{72.00} & \hcb{86.43} & \hc{67.50} & \hc{46.25} & \hc{68.75} & \hc{63.12} & \hc{68.00} & \hcb{66.18} \\
\hline
\end{longtable}
}

\begin{figure}[h]
    \centering
    \includegraphics[width=1\linewidth]{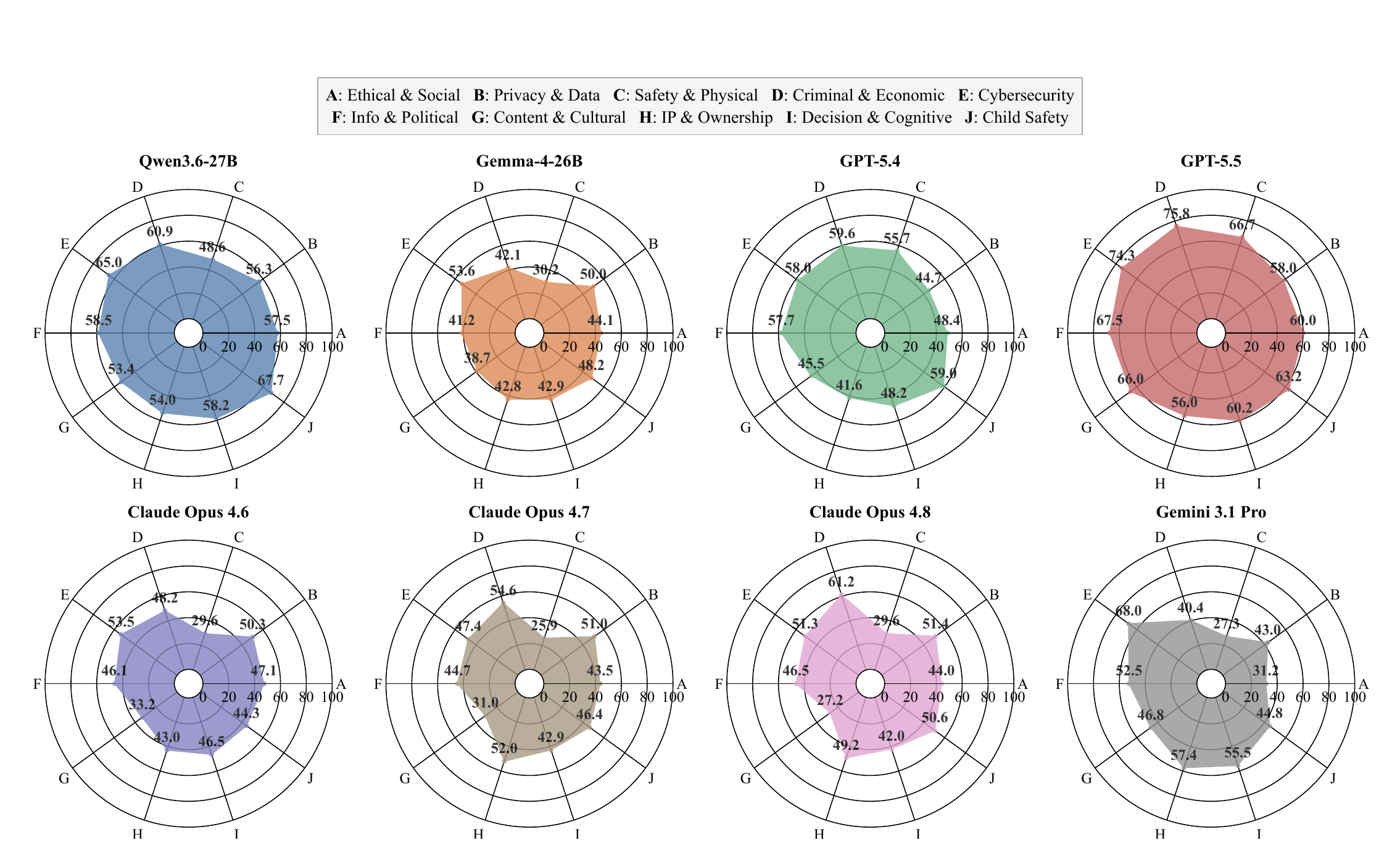}
    \caption{\small Examining model vulnerability against harmful categories}
    \label{fig: selected_avg_vol_corpus}
\end{figure}

While the weakness of a model with respect to a given attack gives an interesting insight in how to chose the attack strategy, one may be interested in global weaknesses of a model. To approximate such information we averaged the results across attack strategies and reported the results in \cref{fig: selected_avg_vol_corpus}. As discussed, the $7\,826$ intents in the PHANTOM benchmark provide sufficient coverage across categories to confidently analyze model vulnerabilities with respect to these specific domains.

% The radar diagrams reveal several critical insights:
% \begin{itemize}
%     \item \textbf{Model Resilience:} With respect to resilience, \textit{Claude Opus 4.6} clearly exhibits the strongest performance, as indicated by the smallest area in the radar plot. In contrast, \textit{GPT‑5.4} and \textit{Gemini 3.1 Pro} cover substantially larger areas, suggesting lower overall resilience.
%     This discrepancy is primarily driven by differences in the frequency of hard refusals (\texttt{null} answers) observed during evaluation. \textit{Claude Opus 4.6} shows a markedly higher propensity to refuse requests outright, whereas \textit{Gemini 3.1 Pro} tends to attempt responses more aggressively. As a result, Gemini more frequently produces partial or unintended harmful content.
%     \item \textbf{High-Risk Domains:} Across the model suite, \textit{F. Info \& Political} and \textit{I. Decision \& Cognitive} appear as common points of vulnerability, particularly for models like \textit{DeepSeek-VL2} and \textit{GLM-4.6V-Flash}, where the ASR reaches its peak.
%     \item \textbf{Category Consistency:} Categories such as \textit{H. IP \& Ownership} and \textit{B. Privacy \& Data} generally show lower ASR scores across the board, suggesting that existing safety alignments are more effective in these domains.
% \end{itemize}

The radar diagrams provide an immediate, at-a-glance understanding of model robustness: a larger colored area corresponds to a greater amount of harmful content produced during testing.
However, an important clarification is needed. In modern models, it is difficult to observe “pure” jailbreaks, i.e., cases in which the model responds directly and fully to a harmful request. Instead, harmful content is more often embedded within longer responses that include benign context and argumentation.
Therefore, these diagrams should be interpreted as indicating a higher tendency of the model to generate harmful content within otherwise complex answers.

With this in mind, models that tend to respond to user requests, even while attempting to avoid harmful content, ultimately produce more harmful content on average. This explains, for example, the stronger performance of Gemma-4-26B compared to many black-box models, which tend to consistently provide answers.
On the other hand, it is important to note that black-box models accessed via API sometimes return \textit{null} responses, most likely due to content filtering mechanisms; we refer to these as \textit{hard refusals}. We treat such cases as failed jailbreaks. Different models exhibit different rates of hard refusals: for instance, the Claude Opus models show a much higher rate of hard refusals compared to both GPT and Gemini, whereas GPT models exhibit the lowest rate.
We now highlight a few observations from the results. In \cref{fig: selected_avg_vol_corpus}, the extent of the colored area allows one to infer, with respect to the chosen baseline judge, the relative robustness of the models: a wider area corresponds to less aligned responses. Among white-box models, Gemma-4-26B is clearly the most robust. 
Among black-box models, the picture is different: all models in the Opus family exhibit comparable robustness, which is also similar to that of Gemini 3.1 Pro, while the GPT family appears less robust. However, as noted earlier, this should be interpreted alongside the higher rate of complete responses they produce.
Interestingly, within the GPT family (from 5.4 to 5.5), performance in terms of alignment appears to degrade slightly, although this is again coupled with the absence of hard refusals.
Another interesting observation from \autoref{tbl:asr_x_model_x_cat_corpus} emerges from the distribution of colored cells: the most effective attacks across all models are those that embed harmful text within images. This suggests that model alignment with respect to embedded textual content remains relatively weak, highlighting a persistent vulnerability in multimodal safety mechanisms.  
Finally, across all models, the most vulnerable categories are \textit{D — Criminal and Economic Risks} and \textit{E — Cybersecurity Threats}, which also correspond to domains where one would expect models to provide more actionable and useful responses.

\section{Limitations}
\label{sctn:limitations}
PHANTOM has several limitations. First, our evaluation relies primarily on an automated judge, Abel-24-HarmClassifier, which may introduce false positives and false negatives, particularly for responses that are partially harmful, evasive, or context-dependent. Although automated judging enables large-scale evaluation, it cannot fully replace human assessment.

Second, our reported evaluation is conducted on sampled subsets of attacks rather than on the entire released dataset. While this makes the evaluation computationally feasible, it may underrepresent variability.

Third, currently we focused on three multimodal attack strategies: BAP, IDEATOR, and MML. These methods were selected for their empirical effectiveness and computational feasibility, but they do not exhaust the space of possible adversarial attacks against VLMs.

Finally, the taxonomy and intent collection may inherit biases from the source benchmarks used to construct the dataset.

\section{Ethical Considerations}
\label{sctn:ethical_considerations}

PHANTOM dataset contains adversarial multimodal samples involving harmful and sensitive intents, and is therefore a dual-use resource. The dataset is intended solely for research, robustness evaluation, and the development of defensive guardrails. To reduce misuse risks, we provide content warnings, structured metadata, category labels. Sensitive categories, including child safety and personally harmful content, are included only for safety evaluation and should be handled under appropriate institutional and ethical safeguards.

\section{Discussion and conclusion}
Our empirical analysis highlights several key insights into the behavior of modern VLMs under adversarial conditions.

First, despite advances in safety training, all evaluated systems exhibit non-negligible ASR across multiple categories, confirming that alignment remains fragile in the presence of carefully constructed multimodal inputs.

Second, we observe significant variation across attack strategies. This suggests that different strategies exploit distinct failure modes of VLMs across categories of harmful content. Consequently, evaluating robustness using a single attack family risks underestimating model vulnerability. In particular, attacks that embed harmful requests directly within images (i.e., typographic attacks such as MML, FC Attack, and CSDJ) achieve consistently higher success rates, indicating that even recent large-scale models still struggle to maintain strong filtering capabilities in fully multimodal settings. A particularly surprising result comes from the Gemma-4-26B model: when attacked, it achieves an ASR of 100\% in two categories, highlighting a substantial vulnerability despite the model's overall capabilities.

Third, the results reveal clear evidence of cross-model transferability, particularly from open-source to proprietary systems. Attacks generated in a white-box setting retain their effectiveness when transferred to black-box models, with ASR remaining above 20\% in most cases and reaching peaks of nearly 80\% for the CSDJ attack. This indicates that vulnerabilities are not purely model-specific but instead reflect shared structural or training-induced weaknesses. These findings have important implications for real-world deployment, where adversaries may optimize attacks against accessible models and subsequently transfer them to closed systems.

A further observation is the heterogeneity across risk categories. Certain domains, such as cybersecurity or economic crimes, tend to yield higher attack success rates, reaching up to 90\% on black-box models, while others are more robust. This variability suggests that current alignment procedures may unevenly cover the safety landscape, leaving gaps that adversarial methods can exploit.
These findings are consistent with the broader trend illustrated in the evaluation results.

Importantly, our analysis also highlights a well known evaluation caveat: success rates depend on the chosen judge model and may be influenced by partial refusals due to external filters or ambiguous outputs (see also, \cite{validMeasurement}, \cite{schwinn2026flipcoin}). 
As such, the results should be interpreted as relative indicators of robustness, rather than absolute measures of harmfulness.
Overall, PHANTOM enables a more systematic understanding of how different attack strategies, model architectures, and safety domains interact, offering a baseline to study multimodal robustness.

By consolidating multiple attack strategies and providing structured evaluation across a diverse set of VLMs, the dataset addresses a key gap in the current landscape: the lack of accessible, reproducible, and comprehensive adversarial resources.
Our results demonstrate that multimodal jailbreaks remain a persistent and transferable threat, that robustness varies significantly across both models and safety domains, and that a diverse set of attack strategies is necessary for reliable evaluation. Beyond benchmarking, PHANTOM provides a practical foundation for future research: the dataset can be used to develop and evaluate defensive mechanisms and guardrails, to train adversarially robust models, and to advance the study of cross-modal alignment failures.
We release PHANTOM with the goal of lowering the barrier to multimodal safety research and fostering more reproducible, standardized, and comprehensive evaluations. We hope that this resource will contribute to a deeper understanding of VLM robustness and support the development of safer and more reliable multimodal AI systems.

\newpage
\bibliographystyle{unsrtnat}
\bibliography{refs}  %%% Uncomment this line and comment out the ``thebibliography'' section below to use the external .bib file (using bibtex)

\newpage
\appendix
\crefalias{section}{appendix}
\crefalias{subsection}{appendix}   % if you reference subsections too
\crefname{appendix}{App.}{Apps.}
\Crefname{appendix}{App.}{Apps.}

\section{Categories and subcategories}\label{apdx: categories and subcategories}

\setlength{\tabcolsep}{6pt}
\renewcommand{\arraystretch}{1.15}

% \small
\begin{longtable}{p{1cm} p{3.2cm} p{7.5cm}}
\caption{\small Taxonomy of Risk Categories and Subcategories}
\label{tab:risk-taxonomy} \\
\toprule
\textbf{Code} & \textbf{Category} & \textbf{Subcategories (Code -- Name)} \\
\midrule
\endfirsthead

\toprule
\textbf{Code} & \textbf{Category} & \textbf{Subcategories (Code -- Name)} \\
\midrule
\endhead

\midrule
\endfoot

\bottomrule
\endlastfoot

A &
\multirow[t]{4}{=}{Ethical and Social Risks} &
A1 -- Bias and Discrimination \\
&& A2 -- Insulting or Harassing Speech \\
&& A3 -- Hate Speech \\
&& A4 -- Discriminatory Behavior \\[0.5ex]

B &
\multirow[t]{5}{=}{Privacy and Data Risks} &
B1 -- Personal Data Leakage \\
&& B2 -- Biometric or Health Data Exposure \\
&& B3 -- Doxxing or Tracking \\
&& B4 -- Surveillance Misuse \\
&& B5 -- Training Data Leakage \\[0.5ex]

C &
\multirow[t]{7}{=}{Safety and Physical Harm} &
C1 -- Dangerous Instructions \\
&& C2 -- Self-harm Promotion \\
&& C3 -- Physical Harm \\
&& C4 -- Human Trafficking \\
&& C5 -- Animal Abuse \\
&& C6 -- Drug Guidance \\
&& C7 -- Environmental Damage \\[0.5ex]

D &
\multirow[t]{5}{=}{Criminal and Economic Risks} &
D1 -- Fraud or Scams \\
&& D2 -- Economic Crime \\
&& D3 -- White-Collar Crime \\
&& D4 -- Illegal Activity \\
&& D5 -- Evasion of Law \\[0.5ex]

E &
\multirow[t]{7}{=}{Cybersecurity Threats} &
E1 -- Malware Generation \\
&& E2 -- Exploit Development \\
&& E3 -- Phishing or Ransomware \\
&& E4 -- Jailbreak or Prompt Injection \\
&& E5 -- Cross-modal Attack \\
&& E6 -- Data Poisoning \\
&& E7 -- Unauthorized Model Access \\[0.5ex]

F &
\multirow[t]{6}{=}{Information and Political Manipulation} &
F1 -- Misinformation or Disinformation \\
&& F2 -- Deepfakes or Synthetic Media \\
&& F3 -- Political Propaganda \\
&& F4 -- Election Interference \\
&& F5 -- Public Order Disruption \\
&& F6 -- National Security Threat \\[0.5ex]

G &
\multirow[t]{4}{=}{Content and Cultural Safety} &
G1 -- Pornography or Nudity \\
&& G2 -- Sexual Content \\
&& G3 -- Violence or Gore \\
&& G4 -- Cultural or Religious Offense \\[0.5ex]

H &
\multirow[t]{4}{=}{Intellectual Property and Ownership} &
H1 -- Copyright Infringement \\
&& H2 -- Trademark or Brand Misuse \\
&& H3 -- Proprietary Data Leakage \\
&& H4 -- Unauthorized Style or Persona Imitation \\[0.5ex]

I &
\multirow[t]{8}{=}{Decision and Cognitive Risks} &
I1 -- Medical Advice \\
&& I2 -- Health Misguidance \\
&& I3 -- Legal Advice \\
&& I4 -- Government Decision Support \\
&& I5 -- Financial Advice \\
&& I6 -- Market Manipulation \\
&& I7 -- Psychological Manipulation \\
&& I8 -- Cognitive Bias or Overreliance \\[0.5ex]

J &
\multirow[t]{5}{=}{Child Safety} &
J1 -- CSAM \& Sexualization \\
&& J2 -- Grooming or Enticement of Minors \\
&& J3 -- Child Trafficking \\
&& J4 -- Harmful Content Targeting Minors \\
&& J5 -- Age Verification Evasion \\

\end{longtable}
\section{Language translation analysis of adversarial attacks}\label{apdx: lang_trans_ablation}

\Cref{tbl:translation_heatmap} evaluates how vulnerable different multimodal large language models (MLLMs) are when safety-critical prompts are translated into various languages or presented in mixed-language settings. Current literature suggests that utilizing low-resource languages should increase model vulnerabilities compared to high-resource ones, as safety alignment data is typically scarce in those languages~\cite{yong2023low}. To test this hypothesis, we select Farsi and Turkish as target low-resource languages. Furthermore, utilizing a mixture of languages within a translation should obscure prompt intent, heighten deception, and ultimately increase the ASR~\cite{song2025multilingual}. For this multi-lingual setting, we choose three high-resource languages (Italian, French, and German) and three low-resource languages (Turkish, Farsi, and Khmer), performing a sentence-by-sentence translation of the adversarial text. As a third approach, we target specific semantic segments by translating only the inherently harmful parts of the prompt into a low-resource language (\textit{Partial Turkish}) to isolate its effect on model safety.

% \begin{figure}[H]
%     \centering
%     \includegraphics[width=\linewidth]{images/translation_heatmap.pdf}
%     \caption{\small Examining model vulnerability against language translation}
%     \label{fig: translation}
% \end{figure}

% \small
% \setlength{\tabcolsep}{3pt}

\begin{table}[ht]
\small
\centering
\caption{\small Examining model vulnerability against language translation. ASR (\%) per model across languages.}
\label{tbl:translation_heatmap}
\begin{tabular}{lccccccccc}
\hline
\textbf{Model} &
\textbf{Baseline} &
\textbf{Farsi} &
\textbf{Italian} &
\textbf{German} &
\textbf{Chinese} &
\textbf{Turkish} &
\textbf{\shortstack{Partial\\Turkish}} &
\textbf{\shortstack{Mixed\\Low-Res}} &
\textbf{\shortstack{Mixed\\Multiling.}} \\
\hline
\textbf{Qwen3.6-27B}    & \hc{43.0} & \hc{42.4} & \hcb{46.9} & \hc{42.8} & \hc{29.5} & \hc{45.7} & \hc{42.4} & \hc{23.8} & \hc{40.6} \\
\textbf{Qwen3-VL-30B}   & \hc{53.5} & \hc{58.8} & \hc{60.4} & \hc{62.0} & \hc{51.1} & \hcb{69.1} & \hc{55.8} & \hc{41.8} & \hc{51.1} \\
\textbf{Qwen3.5-27B}    & \hc{43.4} & \hc{49.7} & \hcb{52.1} & \hc{44.4} & \hc{19.4} & \hc{50.7} & \hc{40.6} & \hc{42.8} & \hc{49.5} \\
\textbf{Ministral-3-14B}& \hc{80.2} & \hc{79.4} & \hc{79.6} & \hc{81.8} & \hc{83.6} & \hc{82.8} & \hcb{84.8} & \hc{57.4} & \hc{81.2} \\
\textbf{gemma-4-26B}    & \hc{45.7} & \hc{53.3} & \hc{51.5} & \hc{52.3} & \hcb{62.4} & \hc{57.6} & \hc{53.5} & \hc{50.1} & \hc{50.1} \\
\textbf{DeepSeek-VL2}   & \hcb{60.4} & \hc{10.3} & \hc{36.4} & \hc{42.0} & \hc{40.2} & \hc{14.1} & \hc{32.1} & \hc{6.9} & \hc{42.4} \\
\textbf{GLM-4.6V-Flash} & \hcb{84.4} & \hc{72.9} & \hc{72.9} & \hc{66.9} & \hcb{84.4} & \hc{62.0} & \hc{80.2} & \hc{36.4} & \hc{78.0} \\
\textbf{Kimi-VL-A3B}    & \hcb{67.7} & \hc{29.7} & \hc{40.6} & \hc{46.5} & \hc{47.5} & \hc{21.6} & \hc{55.2} & \hc{13.3} & \hc{43.4} \\
\textbf{LLaVA-v1.6-13b} & \hcb{57.2} & \hc{13.9} & \hc{21.0} & \hc{27.5} & \hc{31.3} & \hc{17.6} & \hc{36.4} & \hc{8.7} & \hc{20.8} \\
\hline
\end{tabular}
\end{table}

Our experimental evaluation yields several key insights:

\paragraph{The Vulnerability Trade-off.} 
The general consensus from our experiments indicates that language translation increases vulnerability only up to the point where it does not compromise the model's fundamental semantic understanding of the attack. Because adversarial strategies often rely on intricate, multi-layered roleplay scenarios or convoluted logic, translation can introduce excessive linguistic ambiguity. When this ambiguity disrupts comprehension—as heavily observed in the \textit{Mixed Low-Res} column—the model fails to grasp the underlying prompt intent and generates irrelevant or benign responses. These are classified as non-jailbreaks by the evaluation judge, leading to a sharp decline in ASR for mixed-language settings.
    
\paragraph{Targeted Susceptibility in Specific Model Families.} 
The Qwen, Ministral, and Gemma families exhibit heightened vulnerability when exposed to low-resource languages or hybrid formatting (\textit{Partial Turkish}). For instance, Gemma-4-26B shows a noticeable increase in ASR from a baseline of $45.7\%$ to $53.3\%$ in Farsi and $57.6\%$ in Turkish. This confirms that low-resource translations successfully exploit gaps in the cross-lingual safety alignment of these architectures.
    
\paragraph{Cross-Lingual Robustness and Transfer Variations.} 
Models such as Ministral-3-14B and GLM-4.6V-Flash maintain consistently high vulnerability scores across nearly all language configurations (with Ministral hovering around $80\%$ ASR). This suggests that adversarial prompt structures transfer seamlessly across linguistic boundaries for these models. Conversely, models like DeepSeek-VL2 and LLaVA-v1.6-13b experience drastic drops in vulnerability when prompts are translated (e.g., DeepSeek-VL2 plunging from a $60.4\%$ baseline to just $10.3\%$ in Farsi). This pattern points to either a brittle multilingual comprehension capability or a defensive posture that defaults to safe rejections when faced with distribution shifts in language.

\section{Transferability Results}\label{apdx: Transferability}
This section presents additional results on the transferability of the attacks in our dataset to a broader set of models, extending those reported in the main corpus.
We follow the same evaluation protocol: for each attack and each subcategory, we sample 20 instances. Once this set is fixed, it is evaluated across a range of different models, enabling a direct comparison within each attack strategy.
Overall, we evaluate these samples on nine white-box models and six black-box models. Due to the large number of generated outputs, we do not perform manual inspection. Instead, we rely on the state-of-the-art judge Able-24-HarmClassifier \cite{harmmetric26}. As a consequence, the reported results should be interpreted as relative to this evaluation baseline rather than as ground truth.

\begin{figure}[H]
    \centering
    \includegraphics[width=1\linewidth]{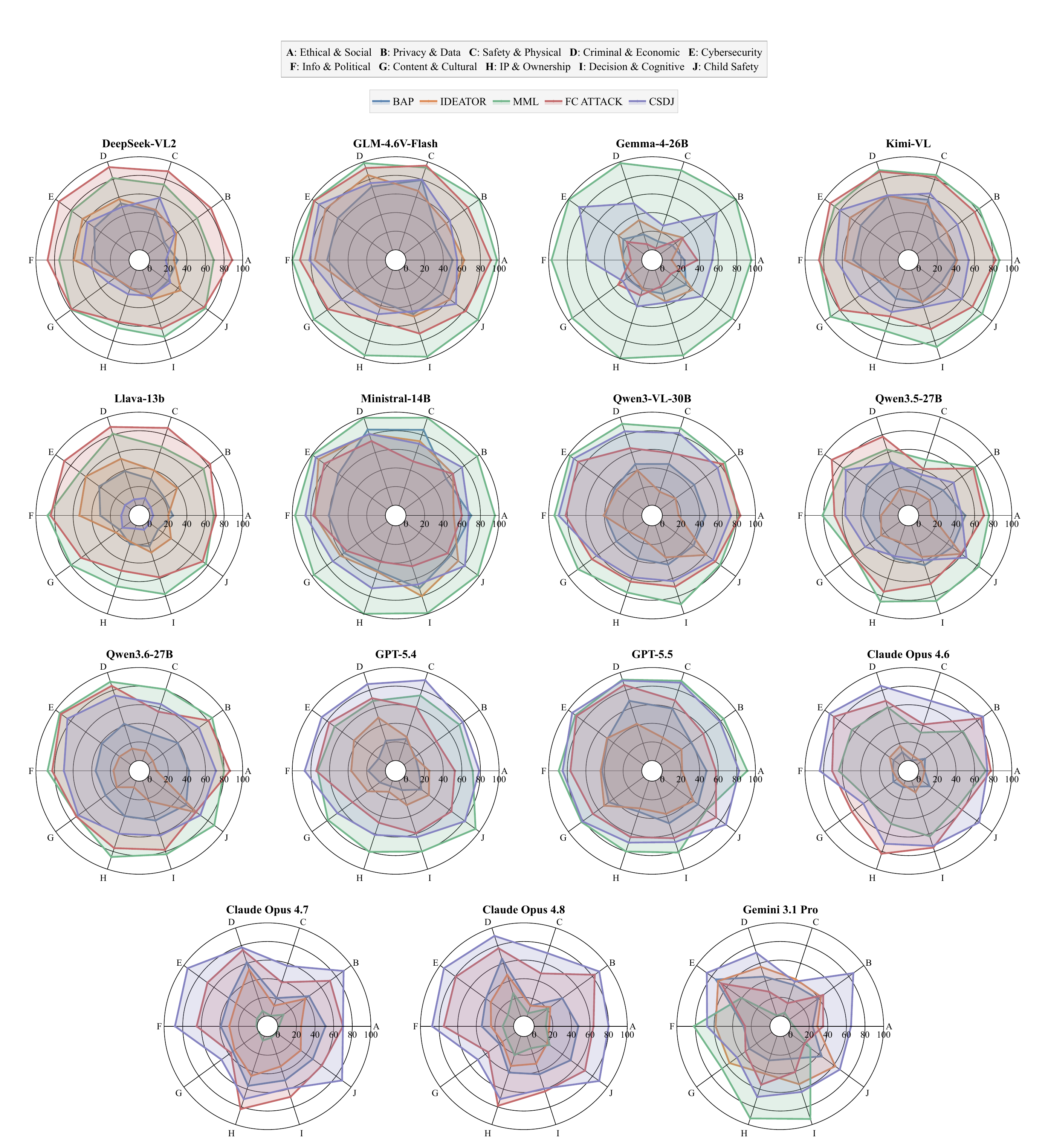}
    \caption{\small The plot graphically presents the vulnerability of each tested model to harmful categories, enabling a comparison across attacks.}
    \label{fig: radar_asr_models_appendix}
\end{figure}

\begin{figure}[H]
    \centering
    \includegraphics[width=1\linewidth]{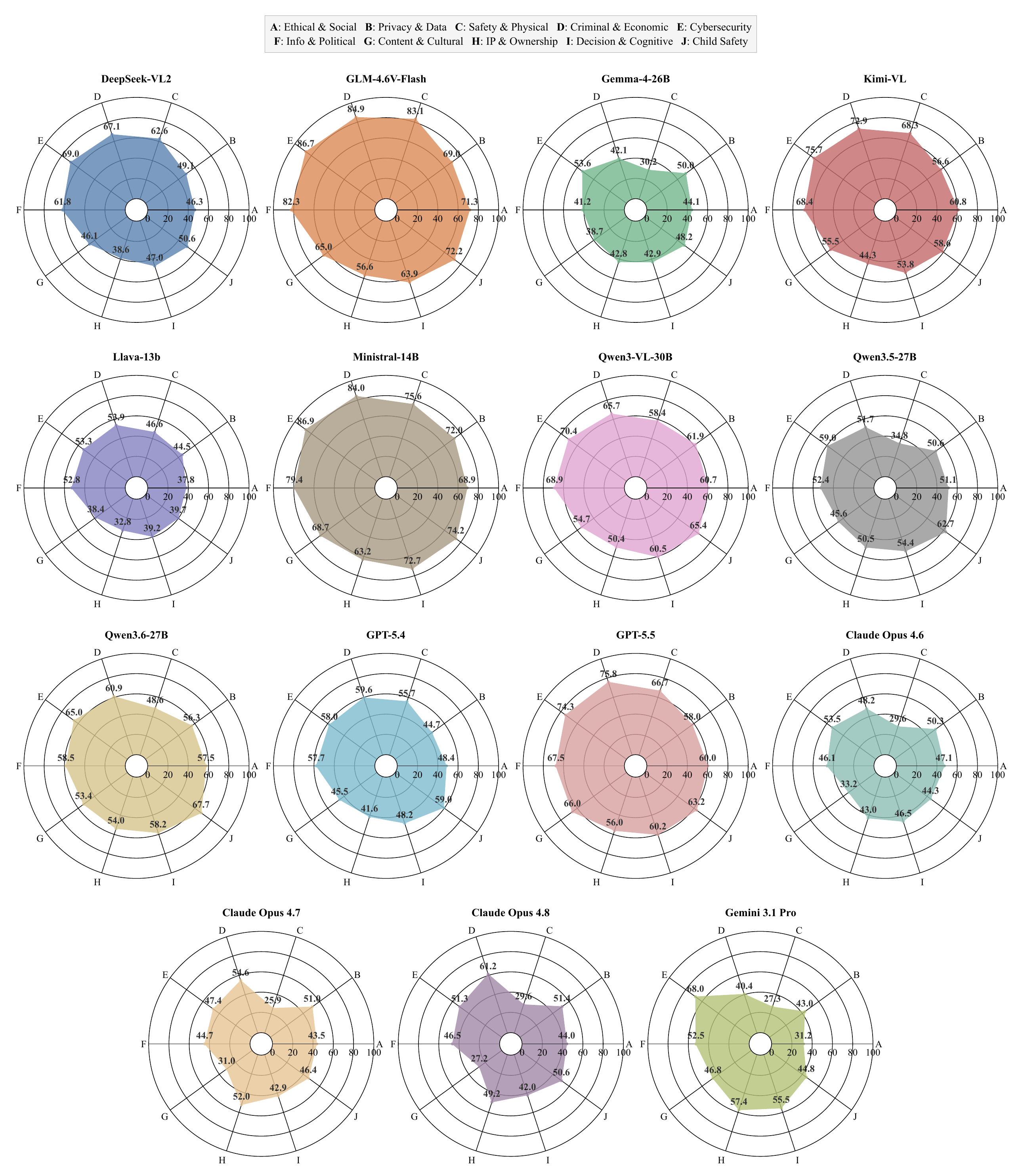}
    \caption{\small The plot presents the vulnerability of models to harmful categories. It is obtained by averaging results across different attack strategies to mitigate the influence of the specific attack used.}
    \label{fig: radar_avg_models_appendix}
\end{figure}

\begin{figure}[H]
    \centering
    \includegraphics[width=1\linewidth]{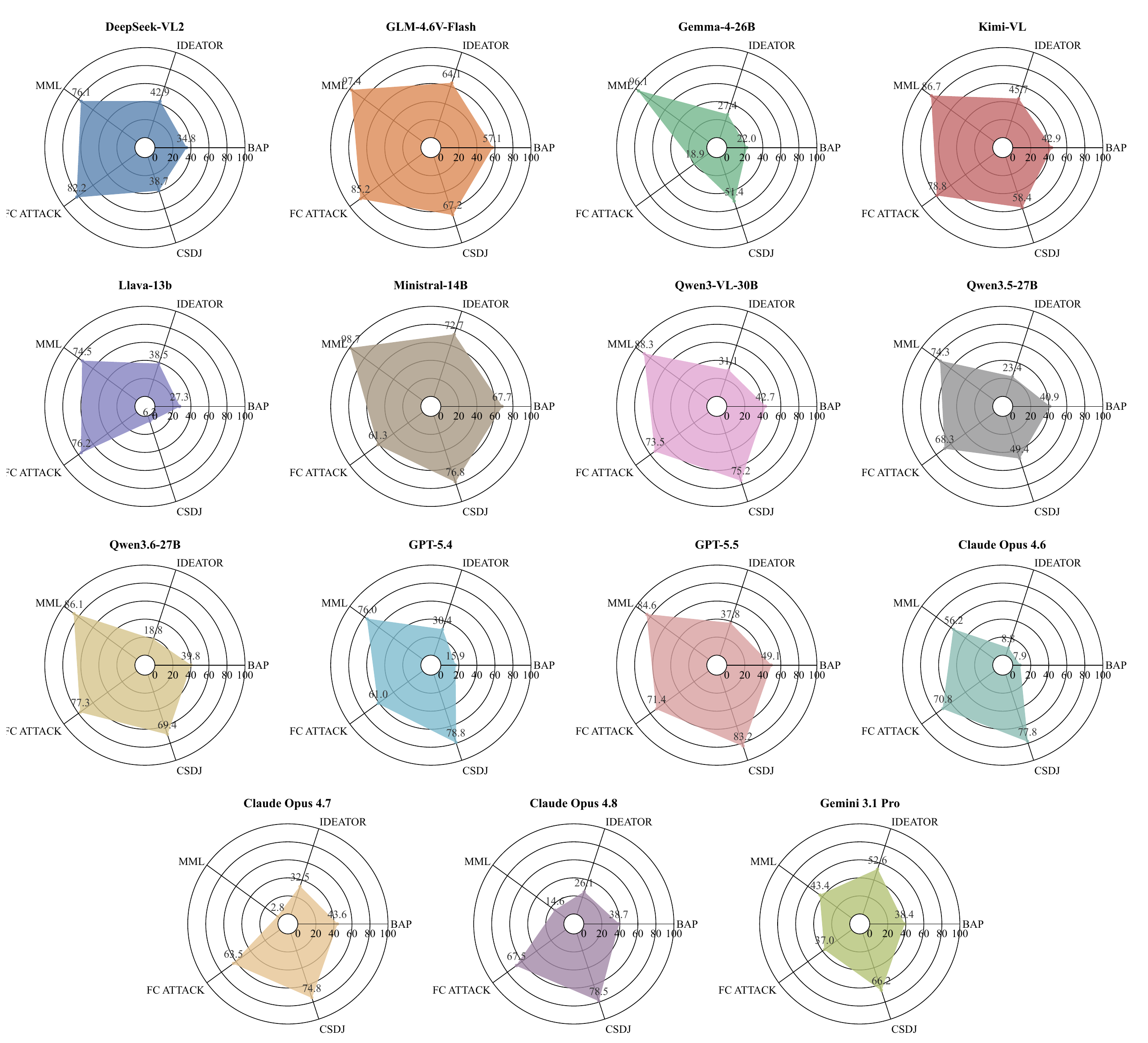}
    \caption{\small The plot shows the vulnerabilities of the models with respect to the attack strategies, averaged over the categories.}
    \label{fig: vul_model_vs_attacks}
\end{figure}

\begin{figure}[H]
    \centering
    \includegraphics[width=1\linewidth]{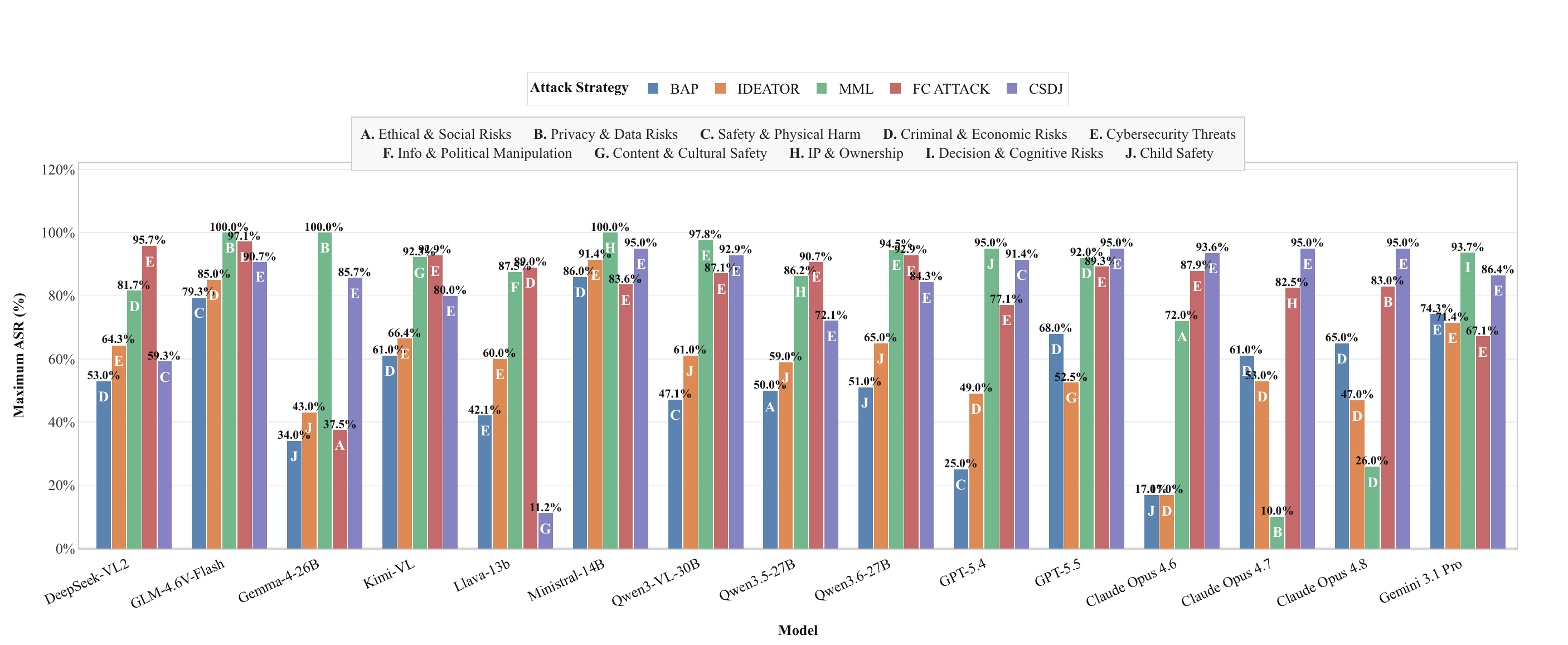}
    \caption{\small This plot identifies, for each attack strategy and model, the most vulnerable category by selecting the category with the highest ASR score.}
    \label{fig: max_vul_model_cat}
\end{figure}

The full results are reported in \cref{tbl:combined-results}. For ease of interpretation, we also provide radar plots offering different insights into the data.
First, \cref{fig: radar_asr_models_appendix} shows the attack success rates across models and attack strategies.
Second, following the analysis in the main paper, \cref{fig: radar_avg_models_appendix} presents the attack success rate (ASR) per category, averaged over attack strategies and evaluated across models, highlighting the categories to which models are most vulnerable independently of the chosen attack.
Third, \cref{fig: vul_model_vs_attacks} shows the ASR averaged over categories, providing an at-a-glance comparison of the most effective attack strategy for each model.
Finally, \cref{fig: max_vul_model_cat} reports the maximum ASR values per model and attack, identifying the weakest category. This allows one to infer, for a given model and attack, the most vulnerable category and the expected performance.

An interesting pattern that emerges from \cref{fig: radar_asr_models_appendix} and \cref{fig: vul_model_vs_attacks} is that MML is the most widely effective attack against almost all models, with the exception of \textit{Opus 4.7} and \textit{Opus 4.8}, which appear to be highly robust to it. However, these two models are particularly vulnerable to the CSDJ attack, which, in turn, is less effective against white-box models.
The second most reliable attack across models is FC Attack, which shows good coverage across categories for most models, except for Gemma-4-26B.
IDEATOR exhibits the most unpredictable behavior: while it achieves high success rates on some white-box models, such as \textit{GLM-4.6V} and \textit{Mistral-14B}, it is generally less reliable, aside from occasional spikes on specific categories.
Finally, BAP yields lower but relatively stable performance across models, with success rates ranging from $30\%$ to $50\%$.

\small
\setlength{\tabcolsep}{3pt}

\begin{longtable}{lccccccccccc}
\caption{\small Table of ASR (\%) per model and per category across all attacks, in bold the category with the highest success rate on each model}
\label{tbl:combined-results}\\

\hline
\textbf{Model} &
\textbf{A} &
\textbf{B} &
\textbf{C} &
\textbf{D} &
\textbf{E} &
\textbf{F} &
\textbf{G} &
\textbf{H} &
\textbf{I} &
\textbf{J} &
\textbf{Average} \\
\hline
\endfirsthead

\hline
\textbf{Model} &
\textbf{A} &
\textbf{B} &
\textbf{C} &
\textbf{D} &
\textbf{E} &
\textbf{F} &
\textbf{G} &
\textbf{H} &
\textbf{I} &
\textbf{J} &
\textbf{Average} \\
\hline
\endhead

\hline
\endfoot

\hline
\endlastfoot

\multicolumn{12}{c}{\textbf{BAP}} \\
\hline
\textbf{DeepSeek-VL2} & 30.00 & 22.00 & 44.29 & \textbf{53.00} & 48.57 & 36.67 & 21.25 & 21.25 & 30.62 & 27.00 & \textbf{34.82} \\
\textbf{GLM-4.6V-Flash} & 50.00 & 43.00 & \textbf{79.29} & 72.00 & 65.71 & \textbf{62.50} & 43.75 & 36.25 & 50.00 & 51.00 & \textbf{57.09} \\
\textbf{Gemma-4-26B} & 23.75 & 16.00 & 10.00 & 18.00 & 27.14 & 20.00 & 22.50 & 22.50 & 26.88 & \textbf{34.00} & \textbf{22.00} \\
\textbf{Kimi-VL} & 40.00 & 29.00 & 57.14 & \textbf{61.00} & 52.14 & 48.33 & 27.50 & 32.50 & 36.88 & 32.00 & \textbf{42.91} \\
\textbf{Llava-13b} & 25.00 & 22.00 & 30.00 & 39.00 & \textbf{42.14} & 30.83 & 15.00 & 21.25 & 23.75 & 14.00 & \textbf{27.27} \\
\textbf{Ministral-14B} & 70.00 & 56.00 & 85.71 & \textbf{86.00} & 65.00 & 60.83 & 57.50 & 51.25 & 71.25 & 62.00 & \textbf{67.73} \\
\textbf{Qwen3-VL-30B} & 46.25 & 45.00 & \textbf{47.14} & 47.00 & 38.57 & 39.17 & 36.25 & 37.50 & 44.38 & 44.00 & \textbf{42.73} \\
\textbf{Qwen3.5-27B} & \textbf{50.00} & 35.00 & 30.71 & \textbf{50.00} & 44.29 & 37.50 & 32.50 & 37.50 & 45.00 & 47.00 & \textbf{40.91} \\
\textbf{Qwen3.6-27B} & 42.50 & 40.00 & 30.71 & 42.00 & 38.57 & 35.83 & 33.75 & 40.00 & 45.00 & \textbf{51.00} & \textbf{39.82} \\
\textbf{GPT-5.4} & 13.75 & 15.00 & \textbf{25.00} & 23.00 & 11.43 & 18.33 & 10.00 & 6.25 & 10.62 & 23.00 & \textbf{15.91} \\
\textbf{GPT-5.5} & 47.50 & 41.00 & 59.29 & \textbf{68.00} & 47.86 & 42.50 & 46.25 & 35.00 & 48.12 & 50.00 & \textbf{49.09} \\
\textbf{Claude Opus 4.6} & 6.25 & 11.00 & 0.71 & 10.00 & 8.57 & 5.83 & 8.75 & 7.50 & 6.88 & \textbf{17.00} & \textbf{7.91} \\
\textbf{Claude Opus 4.7} & 51.25 & 44.00 & 20.71 & \textbf{61.00} & 40.00 & 39.17 & 35.00 & 56.25 & 50.00 & 49.00 & \textbf{43.64} \\
\textbf{Claude Opus 4.8} & 47.50 & 40.00 & 12.14 & \textbf{65.00} & 39.29 & 34.17 & 21.25 & 41.25 & 43.12 & 51.00 & \textbf{38.73} \\
\textbf{Gemini 3.1 Pro} & 23.75 & 39.00 & 35.71 & 45.00 & \textbf{74.29} & 28.33 & 30.00 & 27.50 & 25.62 & 44.00 & \textbf{38.36} \\
\hline
\multicolumn{12}{c}{\textbf{IDEATOR}} \\
\hline
\textbf{DeepSeek-VL2} & 26.25 & 38.00 & 45.71 & \textbf{58.00} & 64.29 & 59.17 & 21.25 & 20.00 & 33.12 & 44.00 & \textbf{42.91} \\
\textbf{GLM-4.6V-Flash} & 62.50 & 56.00 & 67.14 & \textbf{85.00} & 82.14 & 75.83 & 43.75 & 42.50 & 52.50 & 61.00 & \textbf{64.09} \\
\textbf{Gemma-4-26B} & 10.00 & 30.00 & 20.71 & 34.00 & 33.57 & 20.00 & 15.00 & 21.25 & 35.62 & \textbf{43.00} & \textbf{27.36} \\
\textbf{Kimi-VL} & 41.25 & 40.00 & 52.14 & 62.00 & 66.43 & 57.50 & 23.75 & 20.00 & 36.25 & 40.00 & \textbf{45.73} \\
\textbf{Llava-13b} & 20.00 & 39.00 & 40.00 & \textbf{53.00} & 60.00 & 53.33 & 20.00 & 18.75 & 30.62 & 31.00 & \textbf{38.45} \\
\textbf{Ministral-14B} & 51.25 & 65.00 & 72.86 & 81.00 & \textbf{91.43} & 75.83 & 62.50 & 52.50 & \textbf{80.00} & 72.00 & \textbf{72.73} \\
\textbf{Qwen3-VL-30B} & 17.50 & 20.00 & 15.71 & 40.00 & 35.71 & 40.83 & 18.75 & 16.25 & 36.25 & \textbf{61.00} & \textbf{31.09} \\
\textbf{Qwen3.5-27B} & 13.75 & 17.00 & 15.00 & 19.00 & 12.14 & 18.33 & 25.00 & 18.75 & 35.62 & \textbf{59.00} & \textbf{23.45} \\
\textbf{Qwen3.6-27B} & 7.50 & 7.00 & 11.43 & 14.00 & 15.00 & 16.67 & 18.75 & 7.50 & 23.12 & \textbf{65.00} & \textbf{18.82} \\
\textbf{GPT-5.4} & 25.00 & 14.00 & 27.14 & \textbf{49.00} & 44.29 & 36.67 & 26.25 & 12.50 & 27.50 & 33.00 & \textbf{30.45} \\
\textbf{GPT-5.5} & 21.25 & 28.00 & 26.43 & 42.00 & 48.57 & \textbf{44.17} & 52.50 & 31.25 & 37.50 & 44.00 & \textbf{37.82} \\
\textbf{Claude Opus 4.6} & 6.25 & 6.00 & 5.71 & \textbf{17.00} & 13.57 & 5.83 & 3.75 & 5.00 & 13.12 & 7.00 & \textbf{8.82} \\
\textbf{Claude Opus 4.7} & 25.00 & 39.00 & 12.14 & \textbf{53.00} & 30.71 & 30.00 & 32.50 & 45.00 & 34.38 & 33.00 & \textbf{32.55} \\
\textbf{Claude Opus 4.8} & 16.25 & 24.00 & 12.14 & \textbf{47.00} & 32.86 & 24.17 & 17.50 & 33.75 & 31.25 & 20.00 & \textbf{26.09} \\
\textbf{Gemini 3.1 Pro} & 28.75 & 43.00 & 42.14 & \textbf{56.00} & \textbf{71.43} & 58.33 & 56.25 & 42.50 & 54.37 & 62.00 & \textbf{52.64} \\
\hline
\multicolumn{12}{c}{\textbf{MML}} \\
\hline
\textbf{DeepSeek-VL2} & 68.89 & 65.71 & 74.32 & 81.69 & 78.57 & 75.00 & 79.49 & 65.52 & 75.56 & \textbf{76.92} & \textbf{76.10} \\
\textbf{GLM-4.6V-Flash} & 97.78 & \textbf{100.00} & 93.48 & 98.60 & 97.58 & \textbf{100.00} & 97.44 & 96.55 & 98.11 & 98.00 & \textbf{97.36} \\
\textbf{Gemma-4-26B} & 95.56 & \textbf{100.00} & 90.34 & 98.51 & 99.37 & 96.88 & 94.87 & \textbf{100.00} & 96.55 & 95.26 & \textbf{96.09} \\
\textbf{Kimi-VL} & 86.67 & 82.86 & 85.16 & 90.58 & 87.04 & 84.38 & 92.31 & 68.97 & 87.02 & \textbf{87.05} & \textbf{86.66} \\
\textbf{Llava-13b} & 68.89 & 74.29 & 66.48 & 81.34 & 67.92 & \textbf{87.50} & 79.49 & 68.97 & 77.78 & 77.37 & \textbf{74.55} \\
\textbf{Ministral-14B} & 95.56 & 97.14 & 99.43 & 99.25 & 99.37 & 96.88 & 97.44 & \textbf{100.00} & 99.23 & 97.89 & \textbf{98.73} \\
\textbf{Qwen3-VL-30B} & 82.22 & 85.71 & 87.63 & 92.50 & \textbf{97.78} & 93.75 & 87.18 & 75.86 & 89.11 & 79.22 & \textbf{88.35} \\
\textbf{Qwen3.5-27B} & 75.56 & 77.14 & 51.63 & 63.50 & 75.45 & 81.25 & 61.54 & 86.21 & 85.56 & 82.42 & \textbf{74.34} \\
\textbf{Qwen3.6-27B} & 80.00 & 85.71 & 80.97 & 89.63 & \textbf{94.55} & 87.50 & 71.79 & \textbf{86.21} & 83.14 & 88.42 & \textbf{86.10} \\
\textbf{GPT-5.4} & 71.76 & 73.45 & 74.13 & 67.92 & 70.00 & 72.79 & 78.75 & 80.46 & 80.00 & \textbf{95.00} & \textbf{76.03} \\
\textbf{GPT-5.5} & 91.25 & 84.00 & 90.71 & \textbf{92.00} & 90.71 & 89.17 & 82.50 & 80.00 & 81.25 & 61.00 & \textbf{84.64} \\
\textbf{Claude Opus 4.6} & \textbf{71.95} & 61.32 & 32.14 & 60.78 & 63.95 & 63.78 & 41.25 & 48.81 & 62.86 & 54.55 & \textbf{56.19} \\
\textbf{Claude Opus 4.7} & 1.25 & \textbf{10.00} & 0.71 & 6.00 & 2.86 & 0.83 & 0.00 & 5.00 & 1.88 & 1.00 & \textbf{2.82} \\
\textbf{Claude Opus 4.8} & 12.50 & 21.00 & 3.57 & \textbf{26.00} & 10.00 & 11.67 & 11.25 & 21.25 & 13.75 & 23.00 & \textbf{14.64} \\
\textbf{Gemini 3.1 Pro} & 3.53 & 0.88 & 4.20 & 0.94 & 40.62 & 81.62 & 66.25 & 93.10 & \textbf{93.71} & 29.00 & \textbf{43.38} \\
\hline
\multicolumn{12}{c}{\textbf{FC ATTACK}} \\
\hline
\textbf{DeepSeek-VL2} & 88.75 & 84.00 & 89.29 & \textbf{94.00} & 95.71 & 87.50 & 78.75 & 58.75 & 66.25 & 75.00 & \textbf{82.18} \\
\textbf{GLM-4.6V-Flash} & 91.25 & 85.00 & 95.71 & 93.00 & \textbf{97.14} & 91.67 & 78.75 & 57.50 & 71.88 & 82.00 & \textbf{85.18} \\
\textbf{Gemma-4-26B} & \textbf{37.50} & 29.00 & 2.14 & 7.00 & 22.14 & 11.67 & 33.75 & 28.75 & 18.75 & 14.00 & \textbf{18.91} \\
\textbf{Kimi-VL} & 82.50 & 77.00 & 82.86 & 89.00 & \textbf{92.86} & 85.00 & 80.00 & 52.50 & 66.88 & 74.00 & \textbf{78.82} \\
\textbf{Llava-13b} & 71.25 & 83.00 & 87.86 & \textbf{89.00} & 88.57 & 84.17 & 66.25 & 51.25 & 58.75 & 73.00 & \textbf{76.18} \\
\textbf{Ministral-14B} & 60.00 & 65.00 & 50.00 & 73.00 & \textbf{83.57} & 77.50 & 53.75 & 41.25 & 46.25 & 58.00 & \textbf{61.27} \\
\textbf{Qwen3-VL-30B} & 83.75 & 83.00 & 59.29 & 65.00 & \textbf{87.14} & 81.67 & 68.75 & 63.75 & 69.38 & 73.00 & \textbf{73.45} \\
\textbf{Qwen3.5-27B} & 70.00 & 75.00 & 41.43 & 78.00 & \textbf{90.71} & 68.33 & 62.50 & 75.00 & 66.25 & 59.00 & \textbf{68.27} \\
\textbf{Qwen3.6-27B} & 86.25 & 81.00 & 55.71 & 85.00 & \textbf{92.86} & 82.50 & 72.50 & 76.25 & 78.12 & 64.00 & \textbf{77.27} \\
\textbf{GPT-5.4} & 52.50 & 43.00 & 60.71 & 71.00 & \textbf{77.14} & 74.17 & 46.25 & 47.50 & 59.38 & 63.00 & \textbf{61.00} \\
\textbf{GPT-5.5} & 57.50 & 57.00 & 68.57 & 86.00 & \textbf{89.29} & 76.67 & 67.50 & 63.75 & 65.00 & 74.00 & \textbf{71.36} \\
\textbf{Claude Opus 4.6} & 77.50 & 85.00 & 41.43 & 68.00 & \textbf{87.86} & 70.83 & 63.75 & 82.50 & 75.62 & 60.00 & \textbf{70.82} \\
\textbf{Claude Opus 4.7} & 70.00 & 72.00 & 38.57 & \textbf{75.00} & 68.57 & 65.00 & 37.50 & 82.50 & 68.75 & 61.00 & \textbf{63.45} \\
\textbf{Claude Opus 4.8} & 63.75 & 83.00 & 48.57 & 77.00 & \textbf{79.29} & 75.00 & 36.25 & 78.75 & 62.50 & 70.00 & \textbf{67.45} \\
\textbf{Gemini 3.1 Pro} & 35.00 & 46.00 & 15.00 & 28.00 & 67.14 & 26.67 & 35.00 & \textbf{55.00} & 40.62 & 21.00 & \textbf{37.00} \\
\hline
\multicolumn{12}{c}{\textbf{CSDJ}} \\
\hline
\textbf{DeepSeek-VL2} & 17.50 & 36.00 & \textbf{59.29} & 49.00 & 57.86 & 50.83 & 30.00 & 27.50 & 29.38 & 30.00 & \textbf{38.74} \\
\textbf{GLM-4.6V-Flash} & 55.00 & 61.00 & 80.00 & 76.00 & \textbf{90.71} & 81.67 & 61.25 & 50.00 & 46.88 & 69.00 & \textbf{67.15} \\
\textbf{Gemma-4-26B} & 53.75 & 75.00 & 27.86 & 53.00 & \textbf{85.71} & 57.50 & 27.50 & 41.25 & 36.88 & 55.00 & \textbf{51.35} \\
\textbf{Kimi-VL} & 53.75 & 54.00 & 64.29 & 62.00 & \textbf{80.00} & 66.67 & 53.75 & 47.50 & 41.88 & 60.00 & \textbf{58.38} \\
\textbf{Llava-13b} & 3.75 & 4.00 & 8.57 & 7.00 & 7.86 & 8.33 & \textbf{11.25} & 3.75 & 5.00 & 3.00 & \textbf{6.25} \\
\textbf{Ministral-14B} & 67.50 & 77.00 & 70.00 & 81.00 & \textbf{95.00} & 85.83 & 72.50 & 71.25 & 66.88 & 81.00 & \textbf{76.80} \\
\textbf{Qwen3-VL-30B} & 73.75 & 76.00 & 82.14 & 84.00 & \textbf{92.86} & 89.17 & 62.50 & 58.75 & 63.12 & 70.00 & \textbf{75.23} \\
\textbf{Qwen3.5-27B} & 46.25 & 49.00 & 35.00 & 48.00 & \textbf{72.14} & 56.67 & 46.25 & 35.00 & 39.38 & 66.00 & \textbf{49.37} \\
\textbf{Qwen3.6-27B} & 71.25 & 68.00 & 64.29 & 74.00 & \textbf{84.29} & 70.00 & 70.00 & 60.00 & 61.88 & 70.00 & \textbf{69.37} \\
\textbf{GPT-5.4} & 78.75 & 78.00 & \textbf{91.43} & 87.00 & 87.14 & 86.67 & 66.25 & 61.25 & 63.75 & 81.00 & \textbf{78.82} \\
\textbf{GPT-5.5} & 82.50 & 80.00 & 88.57 & 91.00 & \textbf{95.00} & 85.00 & 81.25 & 70.00 & 69.38 & 87.00 & \textbf{83.16} \\
\textbf{Claude Opus 4.6} & 73.75 & 88.00 & 67.86 & 85.00 & \textbf{93.57} & 84.17 & 48.75 & 71.25 & 73.75 & 83.00 & \textbf{77.82} \\
\textbf{Claude Opus 4.7} & 70.00 & 90.00 & 57.14 & 78.00 & \textbf{95.00} & 88.33 & 50.00 & 71.25 & 59.38 & 88.00 & \textbf{74.82} \\
\textbf{Claude Opus 4.8} & 80.00 & 89.00 & 71.43 & 91.00 & \textbf{95.00} & 87.50 & 50.00 & 71.25 & 59.38 & 89.00 & \textbf{78.45} \\
\textbf{Gemini 3.1 Pro} & 65.00 & 86.00 & 39.29 & 72.00 & \textbf{86.43} & 67.50 & 46.25 & 68.75 & 63.12 & 68.00 & \textbf{66.18} \\
\hline
\end{longtable}

\section{Benchmark overview}\label{apdx: benchmark overview}
\subsection{Evolution of Textual and Early Multimodal Benchmarks}
The field was pioneered by \textbf{AdvBench} \cite{zou2023universal}, a text-only benchmark containing $520$ behaviors. Despite its reliance on primitive string-matching judges and the simple GCG attack, it established the foundational framework for subsequent research. Building on this, \textbf{VAJM} \cite{qi2024visual} introduced the image modality and implemented weak categorization based on race and gender. It advanced evaluation methods by utilizing the \textit{Detoxify} classifier as a judge and introducing prompt-tuning optimization attacks.

\textbf{HarmBench} \cite{mazeika2024harmbench} contributed the first rigorous categorization of behaviors into four functional groups. It further matured the evaluation process by introducing a fine-tuned Llama~2 model as a judge and proposing the R2D2 defense mechanism. Expanding the scale of multimodal research, \textbf{JailBreakV-28K} \cite{luo2024jailbreakv} incorporated 16 categories and $2000$ behaviors (sourced from RedTeam-2K), utilizing $28$k attack pairs generated via advanced typographic Stable Diffusion attacks.

\subsection{Diversifying Metrics and Modalities}
Subsequent works focused on refining metrics and interaction types. \textbf{MM-SafetyBench} \cite{mm-safetybench24} introduced the ``Refusal Rate'' as a key metric, emphasizing a comparison between Attack Success Rates (ASR) when models are given text-only queries versus query-relevant image-text pairs. In the purely textual domain, \textbf{Strong Reject} \cite{souly2024strongreject} moved away from binary evaluation by implementing a graded scoring system ($0, 0.33, 0.66, 1.0$) to distinguish between full refusal, partial refusal, partial fulfillment, and full fulfillment across 37 different attacks. 

The importance of conversational context was highlighted by research into \textbf{Multiturn human jailbreaks} \cite{li2024llm}, which demonstrated that models are significantly more vulnerable through iterative, back-and-forth prompting: a feature we leverage through the attack strategy chosen for our dataset. Further expanding the scope of modalities, \textbf{SafeBench} \cite{safebench25} integrated audio alongside text and images, while introducing a ``Safety Index Risk'' evaluated by a consensus-based roundtable of judges rather than a single entity.

\subsection{Balancing Robustness with Model Utility}
A critical shift in the literature involves the trade-off between safety and helpfulness. \textbf{MMJ-bench} \cite{weng2025mmj} categorized attack strategies into optimization- and generation-based methods, arguing that a perfect defense is counterproductive if it causes the model to refuse every prompt. Similarly, \textbf{JailbreakBench} \cite{chao2024jailbreakbench} introduced 100 benign prompts designed to appear harmful but which are actually safe, allowing researchers to measure if a model is overly defensive. To quantify this performance impact, \textbf{B-AVIBench} \cite{zhang2024b} introduced the Average Score Drop Rate (ASDR), measuring the percentage decrease in performance scores following an attack across various image, text, and content bias types.

To ensure automated evaluations remain grounded, \textbf{Sorry-Bench} \cite{xie2024sorry} provided a human validation dataset for judges, utilizing Cohen’s Kappa ($\kappa$) to measure the correlation between AI judges and human evaluation, alongside fulfillment rate as an additional metric.

\subsection{High-Granularity Categorization}
Recent benchmarks have achieved unprecedented depth in their taxonomies. \textbf{VLJailbreakBench} \cite{chao2024jailbreakbench} implemented a robust categorization featuring $12$ safety topics and $46$ subcategories. Finally, \textbf{OmniSafeBench} \cite{omnisafebench25}---which serves as the primary reference for this work, introduced 9 major risk domains and $50$ fine-grained categories. Beyond evaluating $15$ different defense strategies, it established a multifaceted judgment criteria incorporating Harmfulness (H), Intent Alignment (A), and Level of Detail (D) to provide a holistic view of model safety.

 \label{appendix:benchmark_history}
{
\setlength{\tabcolsep}{3pt}
\renewcommand{\arraystretch}{1.05}

\begin{footnotesize}
\begin{longtable}{%
l
p{1.4cm}
c
p{1.6cm}
p{1.2cm}
@{}c c c@{}
p{1.6cm}
l}

\caption{\small Comparative Analysis of Safety and Jailbreak Benchmarks} \label{tab:benchmarks} \\
\toprule
\textbf{Benchmark} & \textbf{Release} & \textbf{Mod.} & \textbf{Behaviours} & \textbf{Samples} & \textbf{Model} & \textbf{Att.} & \textbf{Def. } & \textbf{Judges} & \textbf{Metrics} \\
\midrule
\endfirsthead

\multicolumn{10}{c}%
{{\bfseries \tablename\ \thetable{} -- continued from previous page}} \\
\toprule
\textbf{Benchmark} & \textbf{Release} & \textbf{Mod.} & \textbf{Behaviours} & \textbf{Samples} & \textbf{Model} & \textbf{Att.} & \textbf{Def. } & \textbf{Judges} & \textbf{Metrics} \\
\midrule
\endhead

\bottomrule
\endfoot

OmniSafeBench & 6 Dec 2025 & T/I & $50$ cat. & $1\,200$ & $18$ & 13 & 15 & GPT-4o & ASR, SRI \\
VLJailbreak & 25~Sep~2025 & T/I & $46$ cat. & $3\,654$ & $11$ & 5 & 0 & GPT-4o, GPT-4 & ASR \\
Sorry-Bench & Mar 2025 & T & $44$ Topics & $8\,800$ & $50$ & $0$ & $0$ & Mistral-7B & FR, RR, $\kappa$ \\
AgentHarm & 18~Apr~2025 & T/I & $11$ cat. & $440$ & $15$ & $1$ & 0 & GPT-4o & SR, RR \\
B-AVIBench & 28~Dec~2024 & T/I & $23$ Types & $316$k & $14$ & $10$ & 0 & GPT-4 & ASDR, AED \\
JailbreakBench & 31~Oct~2024 & T & $20$ cat. & $100$ & $4$ & 4 & 0 & 6 Classifiers & ASR, FPR \\
MMJ-Bench & 22~Oct~2024 & T/I & $200$ Behav. & $1\,000$ & $6$ & $9$ & 5 & GPT-4, HarmBench & ASR \\
SafeBench & 4 Oct 2024 & T/I/A & $23$ cat. & $9\,200$ & $21$ & 3 & 0 & Ensemble (2) & ASR, SRI \\
MHJ & 4 Sep 2024 & T & $1\,000$~Req. & $2\,912$ & $1$ & $14$ & $0$ & GPT-4o, HarmBench & ASR \\
StrongREJECT & 27~Aug~2024 & T & $6$ cat. & $346$ & 3 & $37$ & 0 & Gemma 2B & Full Refusal \\
MM-SafetyBench & 19~Jun~24 & T/I & $13$ cat. & $5\,040$ & 12 & 0 & 0 & GPT-4, Llama-2 & ASR, RR \\
JailBreakV-28K & 3 Apr 2024 & T/I & $16$ cat. & $28$k & 10 & 10 & 0 & 4 Classifiers & ASR \\
HarmBench & Feb 2024 & T/I & $4$ cat. & $510$ & $33$ & $22$ & $1$ & Llama-2 (FT) & ASR \\
VAJM & 16~Aug~2023 & T/I & $40$ Behav. & $32\,226$ & $3$ & $1$ & $0$ & Perspective API & Toxicity \\
AdvBench & July 2023 & T & $520$ Behav. & $520$ & $9$ & $8$ & $0$ & GPT-4, String & ASR \\
\bottomrule
\end{longtable}
\end{footnotesize}

\textbf{Legend:} \textit{Mod.} = Modality (T: Text, I: Image, A: Audio); \textit{Model} = Number of models evaluated; \textit{Att.} = Number of attack strategies; \textit{Def.} = Number of defense strategies.

\section{PHANTOM Similarity Checks}

To assess the diversity of the generated adversarial prompts, we performed a cosine-similarity analysis over the textual component of the attacks. This analysis quantifies prompt-level redundancy, including cases where the same underlying intent may lead to multiple generated attacks. Such repetitions are expected, since PHANTOM contains $7\,826$ unique intents but nearly $30$k generated adversarial samples.

We exclude the \texttt{MML} strategy from this analysis because, for this attack, the adversarial content is primarily encoded in the image rather than in the textual prompt. MML and FC ATTACK prompts rely on a shared instruction template, while the harmful intent is embedded through visual transformations such as encoding, mirroring, rotation, or word substitution. Therefore, measuring redundancy using only the textual prompt would produce similarity scores $\sim 100\%$, without providing a meaningful estimate of sample diversity.

\Cref{tab:redundancy_bap_ideator_models} reports the redundancy rates obtained for BAP and IDEATOR across different cosine-similarity thresholds, both globally and broken down by attack strategy and target model. For each threshold $\tau$, we construct clusters of prompts whose pairwise cosine similarity is greater than or equal to $\tau$. Within each cluster, one prompt is treated as the representative, while the remaining prompts are counted as redundant. Formally, if $\mathcal{K}$ denotes the set of clusters and $|C_k|$ the size of cluster $C_k$, the redundancy rate is computed as:
\[
    \text{Redundancy} =
    \frac{\sum_{C_k \in \mathcal{K}} \max(|C_k|-1, 0)}
    {N}
    \times 100,
\]
where $N$ is the total number of prompts in the analyzed group.

\begin{table}[htbp]
\footnotesize
\centering
\caption{\small Redundancy Rate (\%) Across Thresholds}
\label{tab:redundancy_bap_ideator_models}
\begin{tabular}{lccc}
\hline
\textbf{Group} & \textbf{T=80\%} & \textbf{T=85\%} & \textbf{T=90\%} \\
\hline
\textbf{global} (BAP + IDEATOR) & 11.58\% & 9.59\% & 8.91\% \\
\hline
strategy:BAP & 12.10\% & 10.27\% & 9.72\% \\
strategy:IDEATOR & 1.76\% & 0.81\% & 0.22\% \\
\hline
model:DeepSeek-VL22 & 6.45\% & 4.46\% & 3.18\% \\
model:GLM-4.6V-Flash & 9.95\% & 9.67\% & 9.61\% \\
model:Kimi-VL-A3B-Instruct & 7.20\% & 4.34\% & 3.30\% \\
model:Qwen3-VL-30B-A3B & 1.49\% & 0.52\% & 0.38\% \\
model:Qwen3.5-27B & 12.86\% & 12.65\% & 12.55\% \\
model:Qwen3.6-27B & 10.28\% & 8.10\% & 7.70\% \\
\hline
\end{tabular}
\end{table}

\section{Review of the attack strategies}\label{apdx: attack strategies}
In this section we will give an overview of the attack strategies that we used in the generation.

\subsection{BAP attack}

The core idea behind the BAP attack is to jointly optimize visual and textual components. First, an adversarial perturbation is applied to the input image through projected gradient descent (PGD), using a corpus of affirmative model responses as optimization targets. Subsequently, a prompt engineering step is performed to obfuscate the harmful intent within a seemingly benign textual prompt.

The outcome of this process is an adversarial image that biases the model toward affirmative responses, coupled with a carefully engineered prompt that facilitates the bypass of safety mechanisms.

In our pipeline, we fixed the number of PGD optimization steps to approximately 200 and optimized the adversarial image against a batch of 8 affirmative target responses, starting from clean images from the COCO train dataset \cite{cocodataset}. For the prompt engineering phase, we followed the standard iterative interaction flow.

% \[
% \begin{tikzcd}[column sep=large]
% \text{Input}
%   \arrow[r,"\text{forward}"]
% &
% \text{Target}
%   \arrow[r,"\text{evaluation}"]
% &
% \text{Judge}
%   \arrow[r,"\text{feedback}"]
% &
% \text{Attacker}
%   \arrow[ll, bend right=35, "\text{prompt update}" above]
% \end{tikzcd}
% \]

In addition to the target model under attack, we employed an abliterated version of Qwen3.5, namely Huihui-Qwen3.5-9B-abliterated~\cite{huihui_qwen35_9b_abliterated} as the attacker model. Its reasoning capabilities were leveraged via a crafted system prompt that explicitly encoded previous failed attempts. As anticipated before, we used Abel-24-HarmClassifier proposed in~\cite{harmmetric26} as judge model.  

% \textbf{Combined Adaptive PGD System}

% The combined system defines a nested optimization where the adversarial image $x$ and the attack step-size $\alpha$ are updated to maximize the Vision-Language Model's loss $L$. The primary objective is:
% \begin{equation}
%     \max_{x, \alpha} L(\theta, x + \delta(\alpha), t, y)
% \end{equation}

% \textbf{Inner Loop: Image Perturbation}
% The PGD component updates the image at iteration $n$ to maximize model error while remaining within a defined perturbation budget:
% \begin{equation}
%     x_{n+1} = \text{Proj}_{x+\epsilon} \left( x_n + \alpha_k \cdot \text{sign}(\nabla_{x_n} L) \right)
% \end{equation}

% \textbf{Outer Loop: Step-Size Optimization}
% The SGD component updates the attack strength $\alpha$ at meta-iteration $k$ to optimize the efficiency of the adversarial generation:
% \begin{equation}
%     \alpha_{k+1} = \alpha_k - \eta \cdot \nabla_{\alpha} L(\theta, x_{n+1}, t, y)
% \end{equation}

% \textbf{Variable Definitions}
% The following table summarizes the parameters used in the combined adaptive system:

% \begin{table}[h]
% \centering
% \begin{tabular}{ll}
% \toprule
% \textbf{Symbol} & \textbf{Description} \\
% \midrule
% $x_n$ & Adversarial image at step $n$ \\
% $\alpha_k$ & Learnable step-size (perturbation magnitude) \\
% $\eta$ & Meta-learning rate for updating $\alpha$ \\
% $\text{Proj}_{x+\epsilon}$ & Projection onto the $L_\infty$ ball of radius $\epsilon$ \\
% $\nabla_x L$ & Gradient of the loss with respect to image pixels \\
% \bottomrule
% \end{tabular}
% \end{table}

The prompt optimization loop was iterated for up to $5$ attempts for each attack instance.

At first glance, \cref{fig:asr_time} may suggest that this attack is less practical, given that it is substantially slower than the alternatives. However, our decision to include it was motivated by an additional advantage: the adversarial images produced by this pipeline are universal. As a result, one can recombine intents and images to obtain additional valid attacks, although some filtering and discarding may still be required.

\subsection{IDEATOR attack}
Also in the case of this attack strategy, proposed in~\cite{ideator24}, we largely followed the original pipeline. Our modifications mainly consist of introducing different models for prompt and image generation. 

The attack pipeline relies on an attacker that produces two distinct prompts: one used to generate an image related to the harmful intent, and another aimed at engineering the harmful textual prompt itself. Since the process is implemented as a multi-turn conversation in which the prompt is progressively refined, we limited each conversation to a maximum of three image--prompt pairs. In addition, for each target goal, the attack was retried at most three times; these retries serve primarily as a fallback mechanism rather than a core component of the method.

In our implementation, we employed the same ablated Qwen3.5 model above as the attacker to generate both the textual prompts, while image generation was performed using Stable Diffusion 3.5 Medium.

We selected this strategy not only because of its efficiency, but also because its structure naturally supports both multi-turn and single-turn settings: the full conversation can be used as input, or alternatively only the final image--prompt pair can be retained.

\subsection{MML attack}
As with the other methods, we remained faithful to the original structure of the Multi-Modal Linkage attack proposed in~\cite{MMLattack}. In our implementation we start from a harmful or restricted text prompt and apply obfuscation: it replaces key words with benign ones through NLTK package, optionally encodes the text (e.g., Base64), and then renders the transformed text into an image. Additional visual distortions, such as mirroring, rotation, or both, are mainly applied through Pillow library to make the content harder to directly interpret. Alongside this image, the system constructs a carefully designed “game-like” prompt that instructs the model to recover the original text by reversing these transformations (e.g., decoding, un-mirroring, or using a provided word-mapping dictionary) and validating it against a scrambled word list.
The resulting image–text pair is fed into target vision-language model, which is guided step-by-step to reconstruct the original prompt and then generate detailed content based on it. Because the harmful intent is never explicitly presented in raw form but instead reconstructed by the model itself, safety mechanisms can be bypassed. 

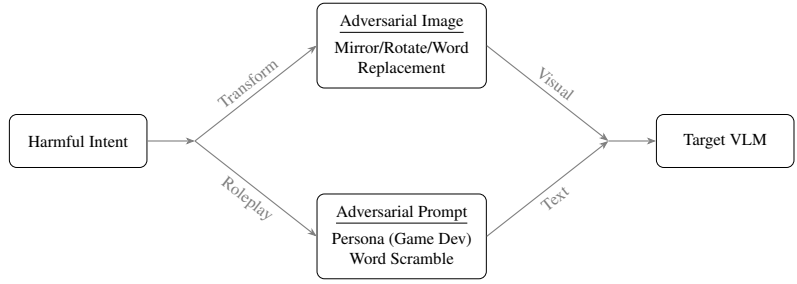
\begin{figure}[H]
\centering
\scalebox{0.7}{
\begin{tikzpicture}[
  box/.style = {
    draw, rectangle, rounded corners=4pt,
    text centered, minimum width=3.2cm, minimum height=1.6cm,
    font=\small
  },
  simplebox/.style = {
    draw, rectangle, rounded corners=4pt,
    text centered, minimum width=2.6cm, minimum height=1.0cm,
    font=\small
  },
  lbl/.style  = {font=\footnotesize, text=gray},
  arr/.style  = {-{Stealth[length=5pt]}, gray, line width=0.6pt},
]

\node[simplebox] (harm) {Harmful Intent};

\node[box, right=3.2cm of harm, yshift=1.8cm] (advimg) {%
  \begin{tabular}{c}
    \underline{Adversarial Image} \\[4pt]
    Mirror/Rotate/Word \\
    Replacement
  \end{tabular}};

\node[box, right=3.2cm of harm, yshift=-1.8cm] (advprompt) {%
  \begin{tabular}{c}
    \underline{Adversarial Prompt} \\[4pt]
    Persona (Game Dev) \\
    Word Scramble
  \end{tabular}};

\node[simplebox, right=3.2cm of advimg, yshift=-1.8cm] (vlm) {Target VLM};

\coordinate (fork)  at ($(harm.east)+(0.9,0)$);
\coordinate (merge) at ($(vlm.west)+(-0.9,0)$);

\draw[arr] (harm.east) -- (fork);
\draw[arr] (fork) -- node[lbl, above, sloped]{Transform} (advimg.west);
\draw[arr] (fork) -- node[lbl, below, sloped]{Roleplay}  (advprompt.west);

\draw[arr] (advimg.east)    -- node[lbl, above, sloped]{Visual} (merge);
\draw[arr] (advprompt.east) -- node[lbl, below, sloped]{Text}   (merge);
\draw[arr] (merge) -- (vlm.west);

\end{tikzpicture}
}
\caption{\small Workflow of MML combining both image manipulation and role-playing through text.}
\label{fig:attack-pipeline}
\end{figure}

\subsection{FC ATTACK}
Once again, our methodology closely follows the original approach proposed in \cite{zhang2025fc}. The core idea is to start from a harmful intent and generate a sequence of logical steps to address it using an auxiliary abliterated model. In our case, similarly to BAP \cite{bap25}, we employ the Huihui-Qwen3.5-9B-abliterated model~\cite{huihui_qwen35_9b_abliterated}. These steps are then represented as a flowchart using standard Python libraries such as \textit{Graphviz}. Finally, the model is prompted with both the flowchart and a standard instruction that encourages it to reason by following the outlined steps.

\subsection{CSDJ attack}
We generated attacks using the CS-DJ attack strategy proposed in \cite{csdj25}.
Following the original pipeline, we crafted each attack as follows.  

We first selected an intent from our dataset and then followed two parallel paths.  
First, using an abliterated model, namely Huihui-gemma-4-31B-it-abliterated~\cite{Huihui-gemma-4-31B-it-abliterated}, we decomposed the harmful request into three less harmful sub-requests and embedded each of them into separate images.  
Second, we selected nine additional images from a pool of 10\,000 images taken from  COCO training set\cite{lin2014microsoft}, these images are chosen such that their CLIP\cite{radford2021learning} embeddings are maximally distant from the embedding of the original intent. 

We then combined these components: the nine images were arranged in a 3×3 grid, followed by the three images containing the generated sub-requests. The images were numbered from 1 (top-left) to 12 (bottom-right).

\end{document}